\documentclass[twocolumn,10pt]{article}

\usepackage{arxiv}
\usepackage[T1]{fontenc}
\usepackage[utf8]{inputenc}
\usepackage{lmodern}
\usepackage{microtype}
\usepackage{hyperref}

\usepackage[american]{babel}
\usepackage[
  backend=biber, 
  natbib=true,
  style=numeric,
  maxcitenames=1,
  mincitenames=1,
  uniquename=false,
  uniquelist=false,
  doi=true,
  isbn=false,
  url=false,
  eprint=false
]{biblatex}
\usepackage{mathtools} 
\usepackage[dvipsnames]{xcolor}
\usepackage{amsfonts}
\usepackage{amsmath}
\usepackage{amssymb}
\usepackage{booktabs} 
\usepackage{makecell}
\usepackage{enumitem}
\usepackage{csquotes}
\usepackage[nonumberlist, toc, section=section]{glossaries}
\usepackage{placeins}
\usepackage{authblk}

\DeclareBibliographyDriver{misc}{%
  \printnames{author}%
  \setunit{\nameyeardelim}%
  \printfield{year}%
  \setunit{\addcolon\space}%
  \printfield{title}%
  \newunit\newblock
  \printfield{journaltitle}%
  \newunit\newblock
  \printfield{url}%
  \finentry
}

\newcommand{\ninit}{N_{\mathrm{init}}}
\newcommand{\nens}{N_{\mathrm{ens}}}
\newcommand{\nmodel}{N_{\mathrm{model}}}

\newcommand{\xhdr}[1]{\textbf{#1}\:}
\renewcommand{\glossarysection}[2][]{}
\makeglossaries

\newglossaryentry{climate}{
    name={Climate}, 
    text={climate}, 
    description={The statistical characterization of \gls{weather}, including averages and variability over extended periods (typically a month or longer), reflecting the slowly evolving components of the coupled atmosphere, ocean (hydrosphere), and land surface system \citep{gettelman2016, AMSclimate}}
}

\newglossaryentry{climate_forecast}{
    name={Climate forecast},
    text={climate forecast}, 
    description={The estimation of future \gls{climate} conditions, focusing on the statistical distribution of variables such as temperature and precipitation over extended periods, typically expressed probabilistically, with \glslink{forecast_lead_time}{lead times} from months to several seasons (see also \gls{climate_model}) \citep{gettelman2016, AMSclimate_prediction}}
}

\newglossaryentry{forecast_lead_time}{
    name={Forecast lead time}, 
    text={forecast lead time}, 
    description={The time interval between the generation of a forecast and the point in time at which the predicted event or condition is expected to occur \citep{AMSlead_time}}
}

\newglossaryentry{climate_model}{
    name={Climate model}, 
    text={climate model}, 
    description={A computational model used to simulate and forecast the behavior of the \gls{climate} system, typically closely related to \glslink{nwp_model}{numerical weather prediction models} but designed for integrations over extended timescales (years to decades or longer) \citep{gettelman2016, AMSclimate_model}}
}

\newglossaryentry{weather}{
    name={Weather}, 
    text={weather}, 
    description={The state of the atmosphere, typically near the Earth's surface, described by variables such as temperature, humidity, precipitation, cloud cover, visibility, and wind. In contrast to \gls{climate}, which characterizes long-term averages and variability, weather refers to short-term atmospheric conditions evolving over timescales from minutes to days 
    \citep{gettelman2016, AMSweather}}
}

\newglossaryentry{nwp_model}{
    name={Numerical weather prediction (NWP) model}, 
    text={numerical weather prediction model}, 
    description={A computational model that simulates the future state of the atmosphere by numerically solving the governing hydrodynamical equations given specified initial conditions for a specific time \citep{gettelman2016, AMSnumerical_forecasting}}
}

\newglossaryentry{weather_forecast}{
    name={Weather forecast}, 
    text={weather forecast}, 
    description={The estimation of the future state of the atmosphere, typically described in terms of variables such as temperature, wind, cloud cover, and precipitation 
    (see also \gls{weather}, \gls{nwp_model}) \citep{gettelman2016, AMSweather_forecast}}
}

\newglossaryentry{ensemble_forecast}{
    name={Ensemble forecast}, 
    text={ensemble forecast}, 
    description={A collection of different \glslink{forecast}{forecasts} that are all valid for the same target time(s) but differ in their inputs or model configurations. 
    The spread among \gls{ensemble} members provides information about uncertainty and can be used to estimate the probability distribution of predicted variables. 
    Ensemble members may differ in initial or boundary conditions, parameter settings, or even the underlying numerical models \citep{AMSensemble_forecast} (see also \gls{ensemble}, \gls{weather_forecast}, \gls{climate_forecast})}
}

\newglossaryentry{ensemble}{
    name={Ensemble}, 
    text={ensemble}, 
    description={A set of model simulations generated with systematic variations, for example in initial conditions, model parameters, or model structure, in order to represent and sample uncertainty \citep{gettelman2016}}
}

\newglossaryentry{model_grid}{
  name={Grid},
  text={grid},
  description={An organized set of points or cells on which model variables are defined, analyzed, and predicted. 
  In \glslink{nwp_model}{numerical weather prediction}, grids may be structured or unstructured, with different horizontal and vertical configurations designed to represent the physical system of interest \citep{gettelman2016, AMSgrid}}
}

\newglossaryentry{model_resolution}{
  name={Model resolution},
  text={model resolution},
  description={The smallest spatial or temporal scale that can be explicitly resolved by a numerical model. 
  It is typically characterized by the size of a \glslink{model_grid}{grid} cell in the horizontal (and sometimes vertical) dimensions
  \citep{gettelman2016, AMSmodel_resolution}}
}

\newglossaryentry{subgrid_scale}{
  name={Subgrid-scale process},
  text={subgrid-scale},
  description={Physical processes that occur at spatial or temporal scales smaller than those resolved by the \glslink{model_resolution}{model resolution}, and therefore cannot be explicitly represented in a numerical simulation. 
  These processes are typically represented through \glslink{parameterization}{parameterizations} or even omitted in some applications \citep{AMSsubgrid_scale_process}.
  In the two-scale Lorenz '96 model, the small-scale $Y$ variables represent \gls{unresolved} (subgrid-scale) processes}
}

\newglossaryentry{resolved}{
  name={Resolved process},
  text={resolved},
  description={Physical processes that are explicitly represented within a numerical simulation because their characteristic spatial or temporal scales are captured by the \glslink{model_resolution}{model resolution}. 
  These are typically associated with larger-scale and slower processes of the system. 
  In the two-scale Lorenz '96 model, the large-scale $X$ variables represent the resolved processes of the system}
}

\newglossaryentry{unresolved}{
  name={Unresolved process},
  text={unresolved},
  description={Physical processes that occur at scales smaller than those represented by the \glslink{model_resolution}{model resolution} and therefore cannot be explicitly resolved in a numerical simulation (see also \gls{subgrid_scale}). 
  These are often associated with smaller-scale and faster processes. 
  In the two-scale Lorenz '96 system, the small-scale $Y$ variables represent unresolved, or subgrid-scale, processes}
}

\newglossaryentry{forecast}{
  name={Forecast},
  text={forecast},
  description={A prediction or estimate of the future state of a system.
  For example, in meteorology, a forecast describes the future state of the atmosphere in terms of variables such as temperature, wind, cloud cover, and precipitation \citep{gettelman2016, AMSforecast} (see also \gls{weather_forecast}, \gls{climate_forecast})}
}

\newglossaryentry{parameterization}{
  name={Parameterization},
  text={parameterization},
  description={A mathematical representation of the physical effects of a process in terms of simplified parameters, used when the process cannot be explicitly resolved by the model resolution, often because it occurs at \glslink{subgrid_scale}{subgrid scales}) (see also \gls{closure}). 
  Empirical parameterizations are typically derived from observed relationships between inputs and outputs \citep{gettelman2016, AMSparameterization}}
}

\newglossaryentry{closure}{
  name={Closure},
  text={closure},
   description={In this work, the term is used analogously to \gls{parameterization}, as is common in practice, to refer to the representation of \gls{unresolved} (\glslink{subgrid_scale}{subgrid-scale}) processes in a numerical model. 
   More precisely, and particularly in turbulence modeling, closure refers to the problem that arises when averaging or filtering the governing equations (e.g., the Navier-Stokes equations), which introduces more unknowns than equations and requires additional assumptions to express the unresolved terms \citep{AMSturbulence_closure}}
}

\newglossaryentry{tendency}{
  name={Tendency},
  text={tendency},
  description={The rate of change of a scalar or vector quantity with respect to time at a given point in space \citep{AMStendency}.
  In the two-scale Lorenz '96 system, tendencies describe how state variables evolve over time.
  For example, \glslink{subgrid_scale}{subgrid-scale} tendencies represent the influence of \gls{unresolved} processes (the fast $Y$ variables) on the time evolution of the \gls{resolved} system (the slow $Y$ variables)}
}

\newglossaryentry{external_forcing}{
    name={External forcing}, 
    text={external forcing}, 
    description={An influence imposed on a system from outside the domain of interest that causes or maintains changes in the system state. 
    The exact definition depends on context; in climate applications, examples include greenhouse gases, anthropogenic aerosols, land-use change, volcanic eruptions, or solar variability. \citep{gettelman2016, AMSexternal_forcing, wills2026}}
}

\newglossaryentry{forced_response}{
    name={Forced response}, 
    text={forced response}, 
    description={The component of climate variability or change that arises in response to \glslink{external_forcing}{external forcing}. 
    It is distinct from unforced internal variability, which includes climate modes such as El Ni{\~n}o-Southern Oscillation (ENSO) and the North Atlantic Oscillation (NAO). 
    Separating forced and unforced components is critical for attributing climate variability and estimating climate change \citep{wills2026}}
}

\newglossaryentry{scenarios}{
    name={Scenarios},
    text={scenarios}, 
    description={Plausible descriptions of future anthropogenic drivers of climate change, such as greenhouse gases, aerosols, and land-use change. 
    In climate modeling, scenarios include forcing pathways to explore possible climate outcomes, as well as socioeconomic development pathways which are used to investigate response options such as mitigation and adaptation \citep{oneill2016, gettelman2016}.
    Because scenarios are plausible conjectures about future developments rather than known boundary conditions, they are an important contributor to uncertainty in climate projections \citep{lehner2020}}
}

\newglossaryentry{spin_up}{
    name={Spin-up time}, 
    text={spin-up time}, 
    description={Initial period of a simulation during which the system adjusts from its initial condition toward its attractor or statistical equilibrium. 
    This transient is usually discarded before analysis. 
    In forced systems, spin-up may also refer to the time required for imposed forcings to take effect \citep{AMSspin_up}}
}

\title{Decomposing Ensemble Spread in Lorenz '96\\ With Learned Stochastic Parameterizations}

\author[1,2,3]{\href{mailto:birgit.kuehbacher@helmholtz-munich.de}{Birgit~K{\"u}hbacher$^*$}}
\author[4,5]{Daan~Crommelin}
\author[1,2,3]{Niki~Kilbertus}

\affil[1]{Technical University of Munich, Munich, Germany}
\affil[2]{Helmholtz Munich, Munich, Germany}
\affil[3]{Munich Center for Machine Learning (MCML), Munich, Germany}
\affil[4]{Centrum Wiskunde \& Informatica (CWI), Amsterdam, Netherlands}
\affil[5]{Korteweg-de Vries Institute for Mathematics, University of Amsterdam, Netherlands}

\begin{document}

\twocolumn[
\begin{@twocolumnfalse}
\maketitle
\end{@twocolumnfalse}
]

\begingroup
\renewcommand\thefootnote{}\footnotetext{* Corresponding author.}
\endgroup

\begin{abstract}
Weather and climate forecasts are inherently uncertain due to chaotic dynamics, imperfect initial conditions, and incomplete representation of the underlying physical processes. 
Operational ensemble forecasts aim to represent these uncertainties through forecast spread, yet many approaches yield underdispersive estimates, with spread that grows too slowly relative to forecast error. 
Using the two-scale Lorenz '96 system as a widely used, controlled testbed, we design a systematic approach to disentangle intrinsic variability, initial-condition perturbations, and stochastic model uncertainty. 
We compare multiple ensemble configurations and parameterization strategies, including existing deterministic and autoregressive as well as novel Bayesian and flow-based approaches. 
Our results show that ensemble perturbations do not increase the system's long-term variance; rather, they regulate how rapidly trajectories decorrelate and explore the invariant measure.
Stochastic parameterizations, particularly those with temporally persistent structure, enhance early spread growth and improve spread-error consistency. 
Overall, we bring clarity to how different sources of uncertainty interact in a chaotic system and provide guidance for the design and evaluation of stochastic parameterizations in weather and climate models.
\end{abstract}

\section{Introduction}
\label{sec:intro}

\glslink{forecast}{Weather and climate predictions} are subject to multiple sources of uncertainty \citep{lehner2020, tebaldi2021, leutbecher2008, palmer2019}. 
While these are conceptually distinct, their relative contributions depend on \glslink{forecast_lead_time}{lead time} and are difficult to separate in practice because they interact dynamically.
This work presents a predictability study that examines how these contributions emerge and grow over the forecast horizon, complementing other uncertainty decomposition studies that focus on long-term climate uncertainty \citep{lehner2020, tebaldi2021, yip2011}, where uncertainty is often characterized by an asymptotic distribution or long-term spread.

We consider three main contributors to the overall uncertainty.
\textbf{(i)}~\emph{Internal variability} arises from the intrinsic chaotic dynamics of the climate system \citep{deser2020, lehner2025}. 
Even with perfect knowledge of the governing aligns and perfect initial conditions, solution trajectories exhibit variability associated with the invariant measure of the dynamical system, i.e., its long-term stationary distribution. 
This variability is irreducible and reflects the natural fluctuations of a nonlinear, chaotic, multiscale system \citep{lehner2020}.
\textbf{(ii)}~\emph{Initial-condition uncertainty} results from imperfect knowledge of the current state of the system \citep{lorenz1963, leutbecher2008}. 
Because the atmosphere and ocean are chaotic, small perturbations in the initial state can grow exponentially, limiting deterministic predictability. 
Unlike internal variability, this source of uncertainty is, in principle, reducible through improved observations and data assimilation, but practical limitations prevent full elimination of initial-condition uncertainty \citep{carrassi2022}.
\textbf{(iii)}~\emph{Model uncertainty} arises from imperfections in numerical models, including structural errors, discretization errors, and imperfect representation of unresolved or insufficiently understood processes \citep{arnold2013, deser2020}. 
In weather and climate prediction, \glslink{parameterization}{parameterizations} of \glslink{subgrid_scale}{subgrid-scale} processes are one of the dominant contributors to model uncertainty, since \gls{unresolved} processes can influence growth and statistical structure of forecast errors \citep{schneider2017, christensen2024}. 

\glslink{ensemble_forecast}{Ensemble} methods are widely used in weather and climate prediction to quantify these uncertainties. 
For long-term climate projections, ensemble spread often reflects differences in \glslink{forced_response}{forced responses}, model formulations, or \glslink{scenarios}{scenario} assumptions \citep{eyring2016, lehner2020, wills2026}, whereas in initial-value prediction the spread is intended to represent the growth of uncertainty with lead time \citep{leutbecher2008, palmer2019}. 
In the latter case, many prediction frameworks remain underdispersive, i.e., their ensemble spread is overly optimistic compared to actual forecast errors \citep{leutbecher2008, berner2017}.
Improving the representation of model uncertainty, particularly through stochastic parameterizations, has therefore become an active area of research \citep{berner2017, leutbecher2017, christensen2024}.
Traditional \glslink{parameterization}{parameterizations} are deterministic, representing the mean impact of unresolved processes, such as cloud formation or deep convection, conditioned on resolved variables. 
Yet, due to incomplete scale separation in the climate system, \gls{unresolved} \glslink{tendency}{tendencies} are generally not uniquely determined by the resolved state \citep{gottwald2017, christensen2024}. 
Stochastic parameterizations address this limitation by modeling a conditional distribution of possible unresolved tendencies and sampling it \citep{berner2017}. 

Studying how these uncertainties affect forecast spread requires controlled experiments in which the different sources can be varied independently and evaluated against a known reference system. 
Because this is difficult to achieve systematically in comprehensive weather and climate models, simplified dynamical systems are often used for predictability and parameterization studies (e.g., \cite{wilks2005, crommelin2008, christensen2015, crommelin2021, gagneii2020, parthipan2023}).
The two-scale Lorenz '96 (L96) system \citep{lorenz2006} is a standard test bed for this purpose, since it combines chaotic resolved dynamics with unresolved fast-scale variability.
We use this system to study forecast uncertainty in a controlled setting, representing model uncertainty through parameterizations of the unresolved fast-scale tendencies and comparing its interaction with internal variability and initial-condition uncertainty across multiple \gls{ensemble} configurations.

\textbf{Our contributions are threefold:}
\textbf{(i)} We rigorously define the different sources of uncertainty and express their contributions through corresponding variance components. 
\textbf{(ii)} We systematically compare deterministic, autoregressive, Bayesian, and novel ML-based stochastic parameterizations within a unified setup. 
\textbf{(iii)} We evaluate and quantify how these formulations influence spread growth, forecast skill, and ensemble calibration using both stationary and dynamical diagnostics.
Overall, our approach clarifies how different sources of uncertainty influence ensemble behavior and provides practical guidance for stochastic parameterizations in weather and climate models.

\section{Background \& Problem Setting}

\subsection{Background}

\xhdr{Ensemble prediction and forecast uncertainty.}
\glslink{nwp_model}{Numerical weather prediction model}s and the atmosphere itself can be viewed as nonlinear dynamical systems whose evolution is sensitive to initial conditions \citep{lorenz1963, leutbecher2008, palmer2019}. 
In this initial-value perspective, imperfect knowledge of the current state and structural model errors lead to \gls{forecast} errors that grow with \glslink{forecast_lead_time}{lead time}, often in a flow-dependent manner \citep{leutbecher2008, palmer2019}. 
\glslink{ensemble_forecast}{Ensemble forecasting} was developed to quantify this state-dependent uncertainty and to estimate not only the forecast itself but also its likely error. 
A forecast system is considered statistically consistent, and therefore reliable, when ensemble spread matches forecast error; however, operational systems frequently exhibit underdispersion: the spread underestimates the error \citep{leutbecher2008, berner2017}. 
This initial-value view differs from long-term \glslink{climate_forecast}{climate projections}, which are usually framed as a boundary-condition problem and focus on distributions of possible climate states under specified \glslink{external_forcing}{external forcings}. 
Accordingly, ensembles over initial conditions, model formulations, and future emission scenarios are used to estimate relative contributions from internal variability, model differences, and \glslink{scenarios}{scenario} uncertainty at fixed projection horizons \citep{yip2011, lehner2020}.

\xhdr{Stochastic parameterizations.}
Atmospheric models discretize the governing aligns on a \glslink{model_grid}{finite grid}, so processes at scales smaller than the \glslink{model_resolution}{grid resolution} (\glslink{subgrid_scale}{subgrid-scale processes}) are not explicitly resolved. 
Their influence on the \glslink{resolved}{resolved variables} is instead represented through \glslink{parameterization}{parameterizations}, which can be interpreted as statistical model specifications linking resolved states to \glslink{unresolved}{unresolved} \glslink{tendency}{tendencies} \citep{berner2017}.
More formally, let $X$ denote the resolved state and $Y$ the unresolved (subgrid-scale) tendencies. 
Deterministic parameterizations aim to specify the conditional expectation $\mathbb{E}[Y \mid X]$, while stochastic parameterizations aim to model conditional probability laws $p(Y \mid X)$ \citep{berner2017, christensen2024}.
This probabilistic formulation allows them to capture unresolved variability and aspects of structural uncertainty due to limited scale separation \citep{gottwald2017}.
In ensemble forecasting, such schemes have been shown to improve
reliability and reduce underdispersion, particularly on
medium-range to seasonal time scales \citep{palmer2009, berner2017, christensen2024}. 
Importantly, both theory and numerical experiments indicate that temporally correlated or persistent stochastic forcing is generally required to reproduce realistic ensemble behavior \citep{gottwald2017, christensen2024}.

\xhdr{Machine-learned parameterizations.}
Recent work has explored machine learning for subgrid modeling, primarily in deterministic form, using neural networks or tree-based methods trained on high-resolution simulations \citep{rasp2018, ogorman2018, grundner2022, yuval2020}. 
These approaches can reproduce mean climate statistics and key variability features, but stability and generalization remain active research topics \citep{rasp2020, lin2025}. 
Stochastic data-based parameterizations have also been proposed \citep{crommelin2008}, for example via multi-member ensembles \citep{behrens2025}, conditional Markov-chains \citep{dorrestijn2013}, or probabilistic random forests \citep{miller2025}, yet systematic evaluation of how such schemes contribute to different sources of forecast uncertainty remains limited.

\subsection{Problem Setting}

\xhdr{The two-scale Lorenz '96 model.}
The two-scale Lorenz '96 model is a low-dimensional chaotic system designed as an idealized representation of atmospheric dynamics \citep{lorenz2006}. 
It is widely used as a testbed for developing parameterizations \citep{crommelin2008, gagneii2020, crommelin2021, parthipan2023}, as well as studying model uncertainty \citep{wilks2005, arnold2013} and regime behavior \citep{christensen2015}. 
Its low computational cost enables extensive experimentation and provides a well-defined notion of ``ground truth,'' while retaining key features of multiscale atmospheric dynamics.
The model consists of two interacting layers of variables arranged on a periodic domain. 
Large-scale variables $X_k$ are each coupled to $J$ small-scale variables $Y_j$. 
The governing aligns are given by
\begin{align}
\dot{X}_k
&=
-X_{k-1}(X_{k-2}\!-\!X_{k+1})
\!-\! X_k
\!+\! F
\!-\!
\frac{hc}{b}
\!\!\!\!
\sum_{j=J(k-1)+1}^{kJ}
\!\!\!\!
Y_j,
\\
\dot{Y}_j
&=
- cb\,Y_{j+1}(Y_{j+2}-Y_{j-1})
- c Y_j
+ \frac{hc}{b}X_{\lfloor(j-1)/J\rfloor +1},\nonumber
\end{align}
where $k = 1, \ldots, K$ and $j = 1, \ldots, JK $.
Both sets of variables satisfy cyclic boundary conditions, $X_{k+K}=X_k$ and $Y_{j+JK}=Y_j$, ensuring spatial homogeneity and statistical equivalence across indices for sufficiently long integrations.
The model parameters and values used in this study are summarized in Tab.~\ref{tab:l96}. 
Following the configuration of \cite{arnold2013}, one model time unit (MTU) corresponds approximately to five atmospheric days, as inferred from error-doubling times in L96 and in the atmosphere \citep{lorenz2006}. 
The fully resolved two-scale system is integrated using a fourth-order Runge-Kutta scheme with time step $\Delta t = 0.001$ and output sampling interval $0.005$.

\xhdr{Reduced model setup and parameterizations.}
The corresponding forecast model is constructed under the assumption that only the large-scale variables $X_k$ are \gls{resolved}, while the influence of the small-scale variables $Y_j$ is \glslink{parameterization}{parameterized} in terms of the resolved state. 
The resulting reduced model takes the form
\begin{equation*}
\dot{X}^*_k
=
- X^*_{k-1}(X^*_{k-2}-X^*_{k+1})
- X^*_k
+ F
-
U_p(X^*_k),
\end{equation*}
where $X^*_k$ denotes the \glslink{forecast}{forecast estimate} of the large-scale state and $U_p$ parameterizes the true \glslink{subgrid_scale}{subgrid} \gls{tendency} $\tfrac{hc}{b} \sum_j Y_j$.
The parameterization can be viewed as a statistical model for the conditional behavior of this subgrid tendency given the resolved state, for example, its conditional mean in the deterministic case.  
To train the parameterizations, the target subgrid tendency is diagnosed once from a single long trajectory of the fully resolved two-scale system. 
Rearranging the large-scale align and approximating time derivatives from discrete model output yields
\begin{equation*}
U(t)
=
- X_{k-1}(X_{k-2}-X_{k+1})
- X_k
+ F
- [X_k]_{\Delta t}(t),
\end{equation*}
for the forward difference $[X_k]_{\Delta t}(t) := \frac{X_k(t+\Delta t)-X_k(t)}{\Delta t}$.
The resulting pairs $(X(t), U(t))$ form the dataset used to train parameterization models. By construction, $U(t)$ represents a time-averaged (aggregated) subgrid tendency over the interval $\Delta t$, rather than the instantaneous coupling term. 
This is consistent with how tendencies are diagnosed in numerical models and reflects that parameterizations act on discrete-time resolved states \citep{wilks2005, arnold2013, gagneii2020, parthipan2023}. 
In all simulations, we integrate the reduced model with a second-order Runge-Kutta scheme using time step $\Delta t = 0.005$.

\begin{table}[t]
\centering
\caption{Lorenz '96 parameter settings.}\label{tab:l96}
\begin{tabular}{lll} 
\toprule
Parameter & Symbol & Value \\ 
\midrule 
Number of large-scale variables $X_k$ & $K$ & 8 \\ 
Number of small-scale variables $Y_j$ & $J$ & 32 \\ 
Coupling constant & $h$ & 1 \\ 
Spatial-scale ratio & $b$ & 10 \\
Time-scale ratio & $c$ & 10 \\
Forcing term & $F$ & 20 \\ 
\bottomrule
\end{tabular}
\end{table}

\section{Experiment Methodology}

\begin{figure*}[t]
    \centering
    \includegraphics[width=1\linewidth]{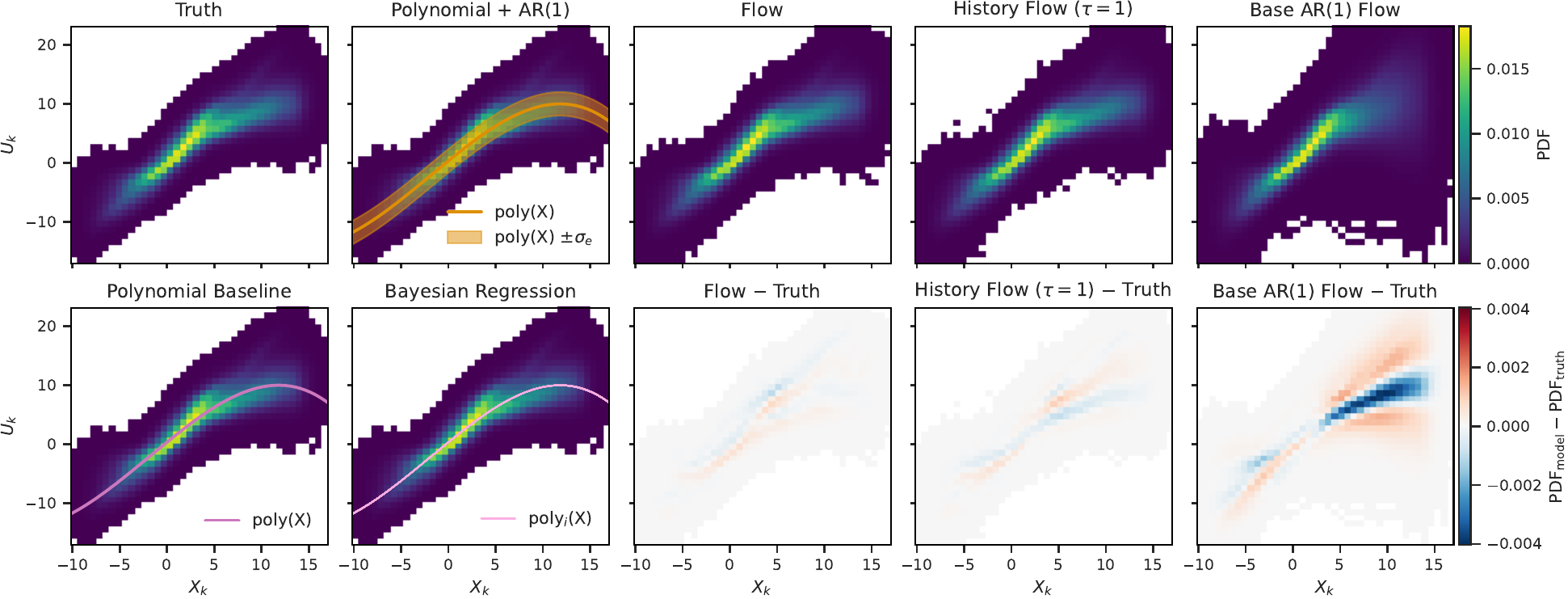}
    \caption{Joint distributions of $X_k$ and $U_k$.
    Deterministic and Bayesian models capture only the mean relationship, resulting in a narrow functional fit (the Bayesian posterior is visually indistinguishable from a line). 
    The polynomial+AR(1) parameterization adds a finite-width corridor via temporally correlated noise. 
    For flow models, $U_k$ is sampled conditionally from test states $X_k$ of the full system. 
    Normal and history flows reproduce the joint structure well, whereas the base AR(1) flow shows larger deviations, reflecting the challenge of learning both the conditional density and latent AR(1) dynamics.}
    \label{fig:xu_fit}
\end{figure*}

\subsection{Parameterization Schemes}
\label{sec:parameterizations}

L96 serves as the reference ``ground truth'' model, while reduced models with parameterized small-scale dynamics are used to study model uncertainty and ensemble spread behavior. 
We consider four classes of parameterizations of increasing flexibility:
\textbf{(i)} deterministic polynomial regression (baseline), 
\textbf{(ii)} polynomial regression with AR(1) residuals, 
\textbf{(iii)} Bayesian polynomial regression, and 
\textbf{(iv)} conditional normalizing flows. 
Fig.~\ref{fig:xu_fit} illustrates the corresponding fits to the true joint $X_k$ and $U_k$ distribution. 
Throughout, $X_k$ denotes a component of the resolved state and $U_k$ the associated subgrid tendency.

\xhdr{Deterministic Polynomial Baseline.}
The deterministic \gls{closure} is modeled component-wise as a cubic polynomial
\begin{equation*}
U_{\mathrm{Det},k}(X_{k,t})
=
aX_{k,t}^3 + bX_{k,t}^2 + cX_{k,t} + d ,
\end{equation*}
where the coefficients $(a,b,c,d)$ are obtained via least-squares regression from training pairs $(X_k,U_k)$.
This parameterization has been used in previous work as a baseline and has been shown to yield a poor forecast model in terms of ensemble spread \citep{wilks2005, crommelin2008, arnold2013}.

\xhdr{Polynomial with AR(1) residual process.}
To represent variability not captured by the deterministic fit, we
augment the cubic closure with an additive stochastic residual,
applied component-wise:
\begin{equation*}
U_{\mathrm{AR1},k}(X_{k,t})
=
U_{\mathrm{Det},k}(X_{k,t})
+
e_{k,t}.
\end{equation*}
The residual is modeled as a first-order autoregressive (AR(1)) process,
\begin{equation*}
e_{k,t}
=
\rho e_{k,t-1}
+
\sigma_e \sqrt{1-\rho^2}\, z_{k,t},
\quad
z_{k,t} \sim \mathcal{N}(0,1), \label{eq:ar1}
\end{equation*}
where $\rho$ denotes the autoregressive coefficient and $\sigma_e$
controls the stationary standard deviation of the stochastic
component.  
The parameters $(\rho,\sigma_e)$ are estimated from the residual time series of the deterministic fit. 
This approach preserves the cubic mean tendency while introducing temporally correlated stochastic forcing to account for subgrid variability \citep{wilks2005, arnold2013, christensen2015}.

\xhdr{Bayesian polynomial regression.}
Rather than estimating a single set of coefficients, we infer a posterior distribution over $(a,b,c,d)$ within a Bayesian linear regression framework. 
The data model assumes
\begin{equation*}
u_i
=
a x_i^3 + b x_i^2 + c x_i + d + \varepsilon_i,
\qquad
\varepsilon_i \sim \mathcal{N}(0,\sigma^2),
\end{equation*}
with weakly informative priors on the coefficients and $\sigma$.
Posterior samples are obtained using the No-U-Turn Sampler (NUTS)
\citep{hoffman2014} as implemented in PyMC \citep{abril-pla2023}.
Each draw defines a deterministic cubic closure, and ensemble variability arises from sampling across coefficient realizations.
Thus, this approach captures \emph{parametric uncertainty} rather than intrinsic stochastic variability. 
As shown in Fig.~\ref{fig:xu_fit}, the posterior concentrates around a point estimate in the high-data regime, causing this uncertainty to vanish. 
Consequently, Bayesian coefficient sampling cannot represent stochastic variability arising from non-deterministic scale interactions, which persists even with arbitrarily large amounts of data. 
We nevertheless include this formulation to test whether uncertainty in the fitted closure coefficients alone can generate meaningful forecast spread.

\xhdr{Conditional normalizing flows.}
Unlike the previous approaches, which prescribe a parametric mean
structure with optional additive noise, we now model the full conditional distribution.
Specifically, we approximate
$p\,(\mathbf{U}_t \mid \mathbf{X}_t)$
using a normalizing flow of RealNVP type \citep{dinh2017}, extended for conditional density estimation \citep{papamakarios2017, trippe2018}.
Let $\mathbf{Z}_t \sim p_0 = \mathcal{N}(\mathbf{0},\mathbf{I})$
denote a latent variable.
The flow defines an invertible transformation
$\mathbf{U}_t = \mathbf{g}_\theta(\mathbf{Z}_t,\mathbf{c}_t),$
where $\mathbf{c}_t$ is a conditioning vector constructed from
resolved variables $\mathbf{X}_t$.
The conditional log density is obtained by change of variables,
\begin{equation*}
\log p_\theta(\mathbf{U}_t \mid \mathbf{c}_t)
=
\log p_0(\mathbf{Z}_t)
+
\log \left|
\det
\frac{\partial g_\theta^{-1}}{\partial \mathbf{U}_t}
\right|.
\end{equation*}
The transformation $g_\theta$ is implemented as a composition of affine coupling layers \citep{dinh2017}, whose scale and shift functions are neural networks that depend on $\mathbf{c}_t$.
Training is performed by conditional maximum likelihood.

\xhdr{Latent temporal correlation.}
After training, latent variables can be sampled either independently,
$\mathbf{Z}_t \sim \mathcal{N}(\mathbf{0},\mathbf{I})$, or evolved
according to an AR(1) process,
\begin{equation*}
\mathbf{Z}_t
=
\rho\,\mathbf{Z}_{t-1}
+
\sigma_z \sqrt{1-\rho^2}\,\boldsymbol{\varepsilon}_t,
\qquad
\boldsymbol{\varepsilon}_t \sim \mathcal{N}(\mathbf{0},\mathbf{I}), 
\end{equation*}
which introduces temporal dependence in latent space. 
The parameters $\rho$ and $\sigma_z$ are estimated from the inferred latent time series. 
Because the resulting stationary marginal distribution need not coincide with the standard normal latent prior used during training, this procedure can introduce a mismatch between training and sampling distributions.
Depending on the flow type, this mismatch may affect forecast performance, which we evaluate empirically.

\xhdr{Flow variants.}
We consider three different architectures: 
\begin{enumerate}[label=\textit{(\roman*)}, itemsep=0.1em, topsep=0.0em, partopsep=0pt, parsep=0pt]
    \item \textit{Normal flow.} Markovian conditioning with $\mathbf{c}_t = \mathbf{X}_t$.

    \item \textit{History flow.} Finite-memory conditioning,
    $
    \mathbf{c}_t
    =
    \big[
    \mathbf{X}_t,
    \mathbf{X}_{t-1},
    \ldots,
    \mathbf{X}_{t-\tau}
    \big],
    $
    which introduces temporal dependence of length $\tau$ through resolved-state history.

    \item \textit{AR(1) base flow.} Instead of imposing temporal correlation only at sampling time, we define the base distribution itself as an AR(1) process and train under the corresponding sequence likelihood.
    This couples latent dynamics and flow parameters during training; see Appendix~\ref{sec:flow_variations_appendix} for details.
\end{enumerate}

\subsection{Ensemble construction}

Ensembles are generated by combining variability from perfect initial states, perturbations of those states, and stochastic model realizations.

\xhdr{Perfect initial states.}
Perfect initial states are extracted from a single long integration $\mathbf{X}^{\mathrm{true}}$ of the fully resolved two-scale L96 system after discarding \glslink{spin_up}{spin-up}, ensuring sampling on the attractor.
We subsample the trajectory at fixed intervals,
\begin{equation*}
\mathbf{X}_{i}(0)
=
\mathbf{X}^{\mathrm{true}}(t_i),
\qquad
i=1,\ldots,\ninit,
\end{equation*}
with constant spacing
$t_{i+1}-t_i=\Delta t_{\mathrm{sel}}=20$ MTU.
This ensures approximate dynamical decorrelation, so that the selected states represent independent samples from the invariant measure \citep{arnold2013}.

\xhdr{Initial-condition perturbations.}
For ensemble members, each perfect state is perturbed with isotropic Gaussian noise
\begin{equation*}
\mathbf{X}_{i,j}(0)
=
\mathbf{X}_{i}
+
\boldsymbol{\eta}_{i,j},
\quad
\boldsymbol{\eta}_{i,j}
\sim
\mathcal{N}\!\left(
\mathbf{0},
(0.05 \cdot \sigma_{\mathrm{clim}})^2 \mathbf{I}
\right),
\end{equation*}
for $j=1,\ldots,\nens$,
where $\sigma_{\mathrm{clim}}$ denotes the climatological standard deviation from a long truth integration.
Following \citet{wilks2005}, the perturbation amplitude is set to
$5\%$ of the climatological variability, ensuring a small but dynamically relevant initial spread. 
In our L96 configuration, $\sigma_{\mathrm{clim}} = 5.07$, resulting in a perturbation standard deviation of $0.25$.

\xhdr{Stochastic model realizations.}
For stochastic parameterizations, additional variability is introduced
through independent stochastic realizations.
Given an initial state $\mathbf{X}_{i,j}(0)$, the reduced model is integrated with different random seeds, or, in the Bayesian regression case, with different draws of the regression coefficients,
$m=1,\ldots,\nmodel$.
This separates variability due to initial-condition perturbations
from variability generated by stochastic physics.

Throughout, we use
$\ninit=300, \nens=20, \nmodel=20$.
Thus, an ensemble configuration $(\ninit \! \times \! \nens)$ corresponds to
$\nens$ perturbed realizations for each of the $\ninit$ initial states.
If independent seeds are also assigned to each member, as in many
operational systems, the two sources of variability cannot be disentangled (see Appendix~\ref{sec:mixed_ensemble_appendix}).
To explicitly separate them, we construct ensembles 
$(\ninit \! \times \! \nens \! \times \! \nmodel)$,
where each perturbed initial state is integrated $\nmodel$ times with independent stochastic realizations.
This configuration enables attribution of spread to perturbation- and model-induced contributions.

\section{Disentangling Uncertainty}
\label{sec:uncertainty_decomposition}

We now formally quantify internal variability, initial-condition uncertainty, and model uncertainty in ensemble forecasts using ensembles indexed by initial state $i=1,\ldots,\ninit$, perturbation $j=1,\ldots,\nens$, and model realization $m=1,\ldots,\nmodel$.

\xhdr{Internal variability.}
Internal variability, reflecting intrinsic fluctuations due to chaotic dynamics, can be interpreted as variability associated with the invariant measure of the system. 
We quantify its amplitude by the climatological standard deviation $\sigma_{\mathrm{clim}}$, a stationary statistic estimated from a long integration ($10\,000$ MTU) and marginalized over time and spatial index $k$. 
Owing to cyclic boundary conditions, all components $X_k$ are statistically equivalent in long integrations.
Furthermore, we characterize variability across attractor-sampled initial states via $\ninit$ short integrations from perfect states and compute
\begin{equation}
V^{\mathrm{init}}_{k}(t)
=
\operatorname{Var}_{i}\!\left(X_{i,k}(t)\right).
\label{eq:var_iv}
\end{equation} 
While $\sigma_{\mathrm{clim}}$ characterizes the typical amplitude of the invariant measure, $V^{\mathrm{init}}_{k}(t)$ provides a finite-sample estimate of variability across attractor-sampled initial states. 
Unlike uncertainty decomposition studies of long-term climate projections \citep[e.g.,][]{yip2011, kirtman2013, lehner2020}, we treat internal variability as a baseline property of the attractor rather than as a variance component in the forecast-spread decomposition.

\xhdr{Uncertainty decomposition.} \label{sec:spread_decomp_aligns}
Using the $(\ninit \! \times \! \nens \! \times \! \nmodel)$ ensemble configuration, we partition ensemble spread into contributions from initial-condition perturbations and stochastic model realizations via an ANOVA-style variance decomposition \citep{smith2014}. 
To isolate the variability generated conditional on a given initial state, we average over $i$, thereby removing variability associated with sampling different locations on the attractor.

The total conditional variance is
\begin{align}
    V^{\mathrm{total},i}_{k}(t)
    =
    \mathbb{E}_{i}
    \left[
    \operatorname{Var}_{j,m}
    \!\left(X_{i,j,m,k}(t)\right)
    \right]. \label{eq:full_var_total_avg_i}
\end{align}

Defining conditional means
\begin{align*}
    \overline{X}_{i,j;k}(t)
    &=
    \mathbb{E}_{m}\!\left[ X_{i,j,m,k}(t) \right], 
    \\
    \overline{X}_{i,m;k}(t)
    &=
    \mathbb{E}_{j}\!\left[ X_{i,j,m,k}(t) \right], 
\end{align*}
the perturbation-induced variance component is
\begin{align}
    V^{\mathrm{ens},i}_{k}(t)
    =
    \mathbb{E}_{i}
    \left[
    \operatorname{Var}_{j}
    \!\left(\overline{X}_{i,j;k}(t)\right)
    \right], \label{eq:full_var_ens}
\end{align}
and the variance induced by stochastic model realizations is
\begin{align}
    V^{\mathrm{model},i}_{k}(t)
    =
    \mathbb{E}_{i}
    \left[
    \operatorname{Var}_{m}
    \!\left(\overline{X}_{i,m;k}(t)\right)
    \right]. \label{eq:full_var_model}
\end{align}
The interaction term between ensemble perturbations and stochastic realizations is defined as
\begin{align}
    I^{\mathrm{ens,model},i}_{k}&(t)
    =
    \mathbb{E}_{i}
    \Big[
    \operatorname{Var}_{j,m}
    \!\left(X_{i,j,m,k}(t)\right)
    \label{eq:full_var_interaction} \\
    -&\operatorname{Var}_{j}
    \!\left(\overline{X}_{i,j;k}(t)\right)
    -
    \operatorname{Var}_{m}
    \!\left(\overline{X}_{i,m;k}(t)\right)
    \Big].\nonumber
\end{align}
This term captures non-additive variability arising from nonlinear coupling between perturbations and stochastic forcing.
In particular, it reflects state-dependent amplification, where the impact of stochastic realizations depends on the perturbed trajectory and vice versa.
By construction, the variance components satisfy
\begin{align*}
V^{\mathrm{total},i}_{k}(t)
=
V^{\mathrm{ens},i}_{k}(t)
+
V^{\mathrm{model},i}_{k}(t)
+
I^{\mathrm{ens,model},i}_{k}(t).
\end{align*}
These quantities express total ensemble variability as the sum of contributions from initial-condition perturbations, stochastic model variability, and their interaction.

\section{Results}
\label{sec:results}
Unless stated otherwise, we report all quantities spatially averaged over $k$ to reduce sampling noise. 
Results for representative single-$k$ components are provided in the Appendix. 
State-space diagnostics are shown without spatial averaging.

\subsection{Sensitivity \& Predictability Baseline}
\label{sec:sensitivity}

\begin{figure}[t]
    \centering
    \includegraphics[width=1\linewidth]{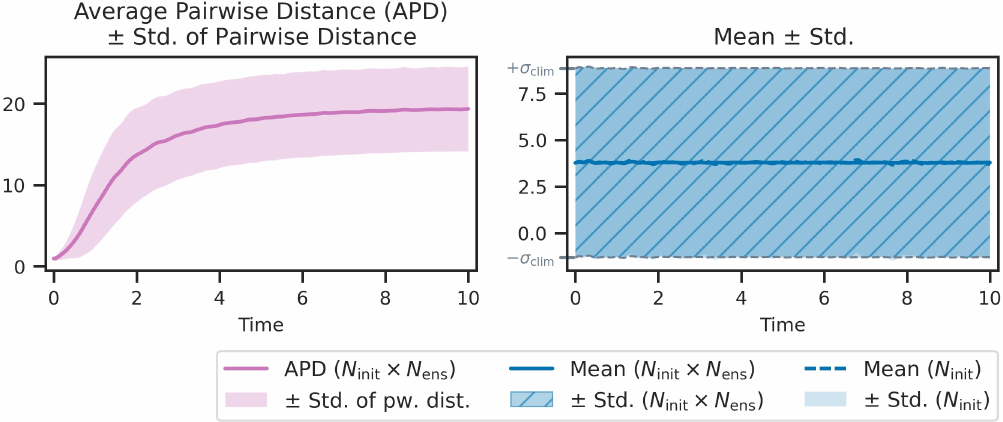}
    \caption{Predictability diagnostics of the fully resolved L96 system.
    Left: growth and saturation of the average pairwise distance.
    Right: ensemble spread relative to $\sigma_{\mathrm{clim}}$.}
    \label{fig:sensitivity}
\end{figure}

We first assess the intrinsic predictability of the fully resolved two-scale L96 system. 
For this purpose, we compare two configurations over 10 MTU: 
(a) $(\ninit \! \times \! \nens)$ integrations initialized from perturbed states, and 
(b) $\ninit$ integrations initialized from perfect attractor-sampled states.
For configuration (a), we compute the average pairwise Euclidean distance (APD) between ensemble members (see Appendix~\ref{sec:sensitivity_appendix} for details). 
The left panel of Fig.~\ref{fig:sensitivity} shows rapid exponential growth of the APD until $t \approx 2$, consistent with chaotic amplification of small initial perturbations, followed by saturation around $t \approx 3$--4 at the attractor scale. 
Thus, initially, nearby trajectories diverge rapidly before becoming distributed across the attractor.
We then compare ensemble means and variances for both configurations. 
For the perturbed ensemble (a), these are computed as
$\mathbb{E}_{i,j}[X_{i,j,k}(t)]$ and 
$\operatorname{Var}_{i,j}(X_{i,j,k}(t))$;
for the perfect-state ensemble (b), they are computed as
$\mathbb{E}_{i}[X_{i,k}(t)]$ and
$\operatorname{Var}_{i}(X_{i,k}(t))$.
The right panel of Fig.~\ref{fig:sensitivity} shows that the means are nearly indistinguishable, suggesting that small perturbations do not affect the large-scale statistics. 
The variances, shown as standard deviations, also largely overlap, indicating that perturbations do not increase the overall ensemble spread beyond the variability already induced by sampling different attractor states. 
Robust summary statistics lead to the same conclusion (not shown; see Fig.~\ref{fig:sensitivity_appendix_avg_k}).
In both configurations, the ensemble spread approaches the climatological standard deviation $\sigma_{\mathrm{clim}}$ at saturation, consistent with the intrinsic variability of the attractor.

\xhdr{Summary.} 
Perturbations mainly control early-time divergence, while long-term variability is bounded by the attractor scale. 
We therefore evaluate reduced models by (i) their ability to reproduce the invariant measure and (ii) spread growth within the predictability window ($t\leq5$).

\subsection{Climatology \& Internal Variability}
\label{sec:climatology}

\begin{figure}[t]
    \centering
    \includegraphics[width=1\linewidth]{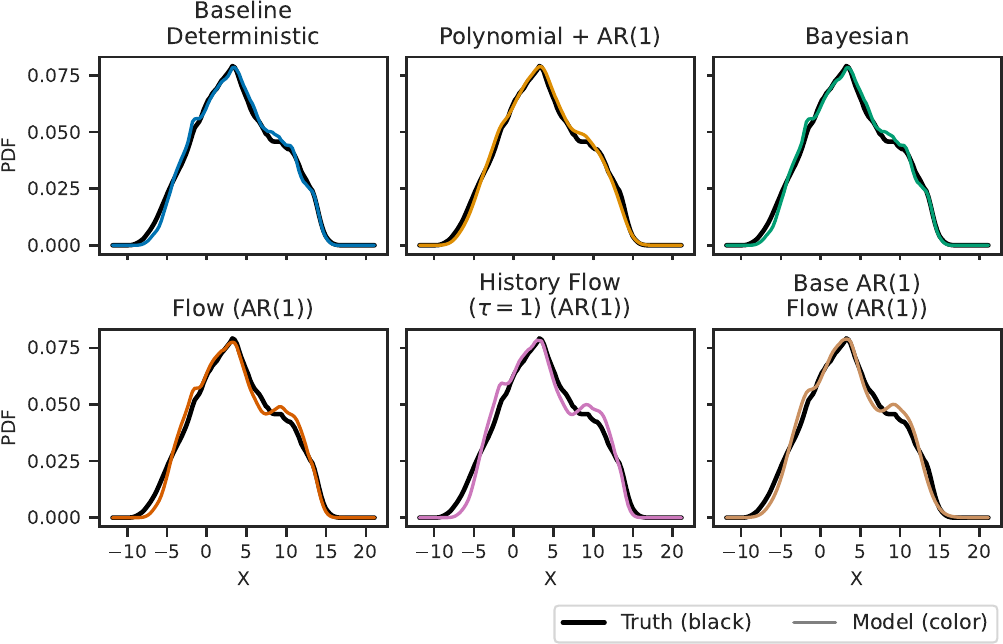}
    \caption{Marginal PDFs of $X_k$ from long integrations.}
    \label{fig:pdf}
\end{figure}

\begin{figure}[t]
    \centering
    \includegraphics[width=1\linewidth]{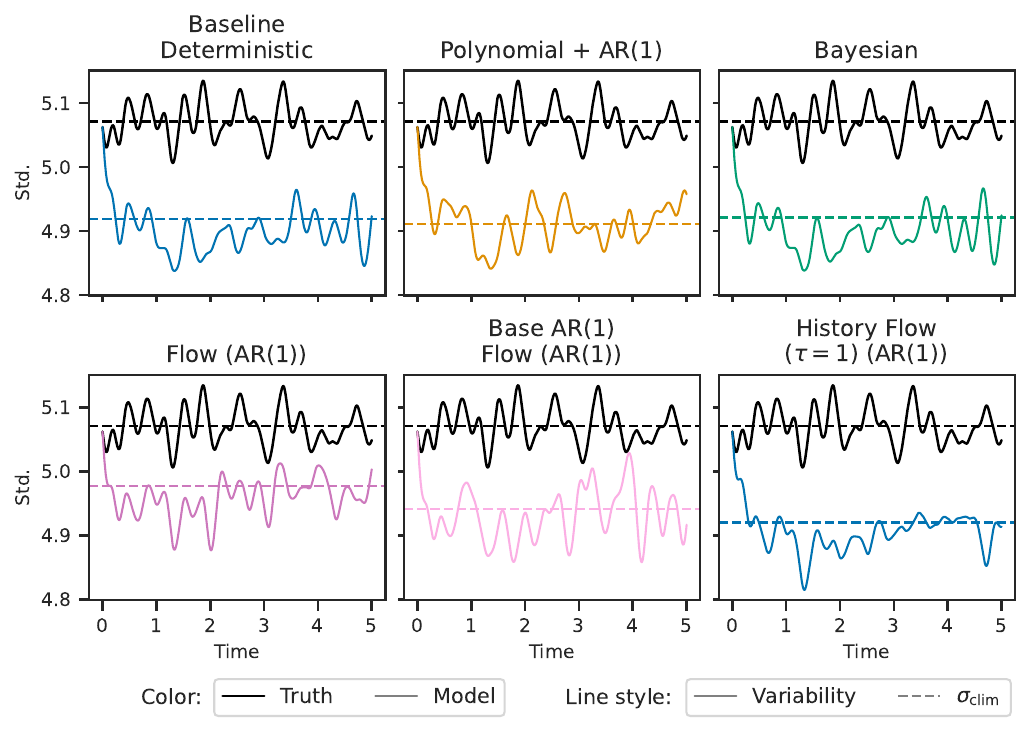}
    \caption{Invariant-measure variability $\sigma^{\mathrm{init}}_k(t)$ and climatological amplitude $\sigma_{\mathrm{clim}}$ for truth and reduced models.}
    \label{fig:iv}
\end{figure}

\begin{table}[t]
\centering
\caption{Hellinger distance to truth for reduced models.}\label{tab:hellinger}
\begin{tabular}{lc}
\toprule
Model & Hellinger Distance \\
\midrule
Baseline Deterministic & 0.0416 \\
Polynomial+AR(1) & 0.0311 \\
Bayesian & 0.0417 \\
Flow (AR(1)) & 0.0570  \\
History Flow ($\tau = 1$) (AR(1)) & 0.0715  \\
Base AR(1) Flow (AR(1)) & 0.0517  \\
\bottomrule
\end{tabular}
\end{table}

\xhdr{Long-term climatology.}  
We next assess whether the reduced models reproduce the long-term distribution of $X_k$, which serves as the analog of climatological statistics in L96.
For both truth and reduced models, we integrate $10$ independent trajectories for $10\,000$ MTU from perfect initial states and aggregate all $X_k$ across all trajectories, time, and space.
Fig.~\ref{fig:pdf} shows the resulting marginal PDFs for selected models, while Tab.~\ref{tab:hellinger} reports the Hellinger distance
\begin{align*}
H(p,q)
=
\frac{1}{2}
\int \left( \sqrt{p(x)}-\sqrt{q(x)} \right)^2\,dx,
\end{align*}
between marginal truth density $p$ and model density $q$ 
(additional Kolmogorov-Smirnov statistics are provided in Appendix~\ref{sec:distributional_metric_appendix}).
All parameterizations reproduce the long-term marginal PDF reasonably well, indicating that the reduced systems remain close to the correct invariant measure. 
The deterministic, polynomial+AR(1), and Bayesian closures show slightly smaller Hellinger distances than the flow variants. 
This reflects the fact that the large-scale attractor of L96 is approximately unimodal and near-Gaussian, so low-order polynomial closures already capture its dominant structure. 
Flow models, by contrast, are trained to approximate the conditional distribution $p(\mathbf{U}_t \mid \mathbf{X}_t)$; small conditional discrepancies can accumulate dynamically and lead to deviations in the stationary marginal. 
Within the flow family, AR(1) sampling improves PDF agreement relative to i.i.d.\ sampling (see Fig.~\ref{fig:pdf_appendix} and Tab.~\ref{tab:hellinger_ks_appendix}), highlighting the importance of temporal correlation for reproducing the invariant measure.
Although generative ML parameterizations can be tuned after training by adjusting the sampling noise magnitude \citep{gagneii2020}, we do not apply such post-hoc calibration here, since it would use evaluation behavior to modify the learned sampling distribution. 
This ensures that all parameterizations are compared based on what they infer from the available training data alone.

\xhdr{Internal variability.} 
We assess the representation of the invariant measure in the reduced models using $\ninit$ trajectories integrated for $t=10$. 
In Fig.~\ref{fig:iv} we show variability defined by the variance $V^{\mathrm{init}}_k(t)$ as the corresponding standard deviation $\sigma^{\mathrm{init}}_k(t)$ to allow direct comparison with $\sigma_{\mathrm{clim}}$. 
All reduced models exhibit a brief adjustment phase, visible as a small initial dip in $\sigma^{\mathrm{init}}_k(t)$, reflecting relaxation from the truth attractor toward their own invariant measure. 
Thereafter, variability fluctuates around the model-specific $\hat{\sigma}_{\mathrm{clim}}$, which is slightly smaller than that of the full system in all cases.
This indicates a modest contraction of the invariant measure. 
Deterministic, polynomial+AR(1), and Bayesian variants yield nearly identical attractor variance, while flow variants show somewhat larger spread across configurations. 
In contrast to the PDF metric, AR(1) versus i.i.d.\ sampling has little influence on $\hat{\sigma}_{\mathrm{clim}}$ (see Fig.~\ref{fig:iv_avg_appendix}), suggesting that temporal correlation primarily affects distributional structure rather than overall amplitude.

\xhdr{Summary.}  
Matching the marginal PDF is a relatively weak test in L96 since attractor statistics are close to Gaussian and well reproduced by simple parameterizations. 
In contrast, small shifts in $\hat{\sigma}_{\mathrm{clim}}$ are structurally important: as the characteristic attractor scale, it sets the upper bound on ensemble spread. 
Reduced attractor variance, therefore, implies a lower spread ceiling and can lead to underdispersion, independent of ensemble design.

\subsection{Spread Decomposition}
\label{sec:spread_decomposition}

\begin{figure}[t!]
    \centering
    \includegraphics[width=1\linewidth]{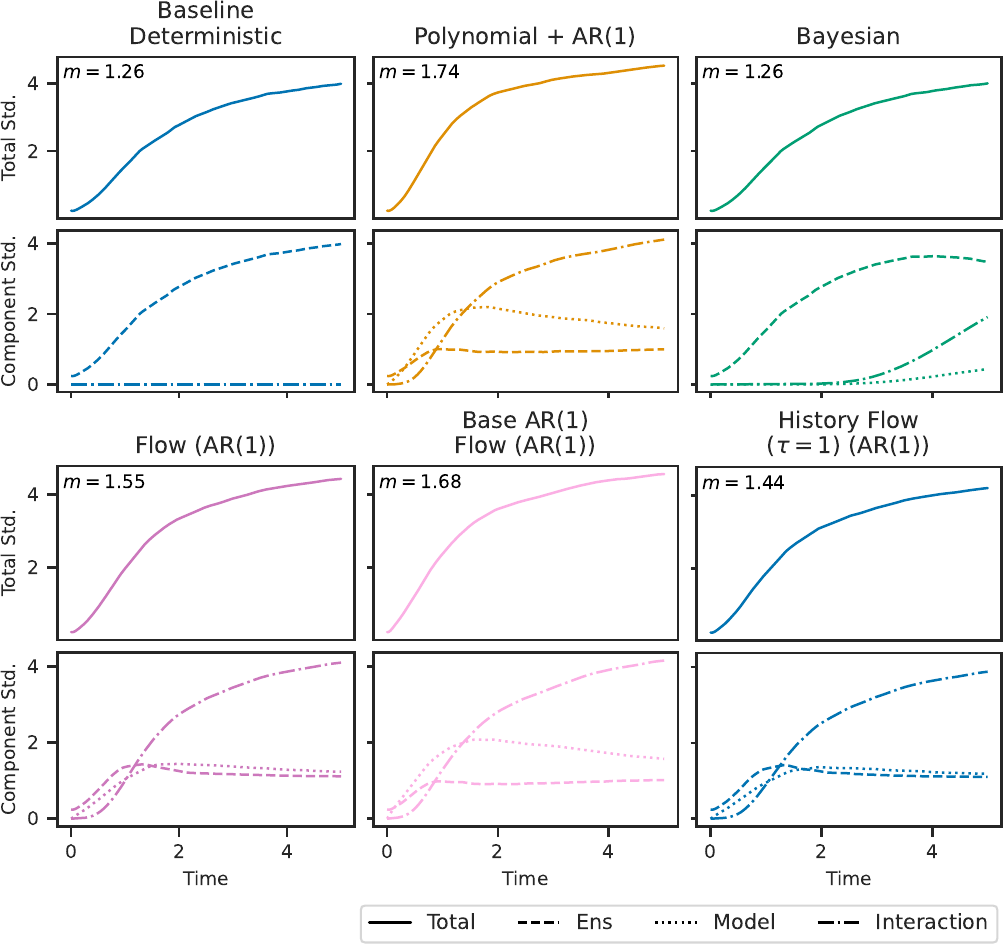}
    \caption{Decomposition of total initial state averaged ensemble spread into perturbation-, model-, and interaction-induced components.
    The value $m$ denotes the early-time growth rate of total spread, computed as the slope of the line from $t=0$ to $t=2$.
    }
    \label{fig:spread_decomposition}
\end{figure}

Using the separated $(\ninit \! \times \! \nens \! \times \! \nmodel)$ configuration, we decompose total spread into perturbation- and model-induced components as described in Eqs.~\eqref{eq:full_var_total_avg_i}-\eqref{eq:full_var_interaction}. 
Fig.~\ref{fig:spread_decomposition} shows the corresponding standard deviations $\sigma^{\mathrm{total},i}_k(t)$, $\sigma^{\mathrm{ens},i}_k(t)$, $\sigma^{\mathrm{model},i}_k(t)$, 
and the interaction term.
As expected, the deterministic baseline shows no model-induced spread. 
For the Bayesian regression, $\sigma^{\mathrm{model},i}_k(t)$ is initially negligible because the posterior over coefficients is narrow; its delayed growth around $t\approx3$ reflects chaotic amplification of small parametric differences. 
Furthermore, whenever the model-induced component is small, early growth is dominated by perturbations, and vice versa. 
Consistent with the sensitivity analysis, adding stochastic variability does not increase the eventual spread level; it alters how quickly the ensemble approaches the attractor-scale variability. 
This rate of early spread growth is quantified in Fig.~\ref{fig:spread_decomposition} by the slope $m$, defined as the linear increase of total spread from $t=0$ to $t=2$.
For most flow variants with i.i.d.\ sampling, this initial growth is largely governed by $\sigma^{\mathrm{ens},i}_k(t)$, whereas $\sigma^{\mathrm{model},i}_k(t)$ increases more gradually (see Fig.~\ref{fig:spread_decomposition_avg_k_appendix}). 
With AR(1) sampling, model-induced variability becomes more comparable to the perturbation component at short lead times, as temporal persistence allows stochastic differences to accumulate coherently. 
The base AR(1) flow is a notable exception: with AR(1) sampling, model-induced spread dominates early growth, closely mirroring the polynomial+AR(1) model. 
In both cases, temporal persistence is embedded directly in the model formulation, so differences between realizations accumulate efficiently, producing the most rapid early increase in total spread.
Moreover, the interaction component is non-zero for all stochastic models, indicating that perturbations and stochastic realizations are dynamically coupled through nonlinear evolution; their contributions are therefore not independent, despite being initialized independently. 
 This interaction becomes dominant after about 1--2 time steps, corresponding to the timescale over which this coupling develops, and grows fastest for stochastic models with temporal persistence.

\xhdr{Summary. } 
Early spread growth is primarily controlled by initial-condition perturbations unless the stochastic forcing exhibits persistence. 
Embedding temporal memory directly in the stochastic parameterization, rather than imposing correlation only at sampling time, most effectively enhances short-range spread. 
Whether this accelerated growth improves reliability is assessed next using forecast skill and calibration diagnostics.

\subsection{Forecast Skill and Calibration}
\label{sec:forecast_skill}

\begin{figure}[t]
    \centering
    \includegraphics[width=1\linewidth]{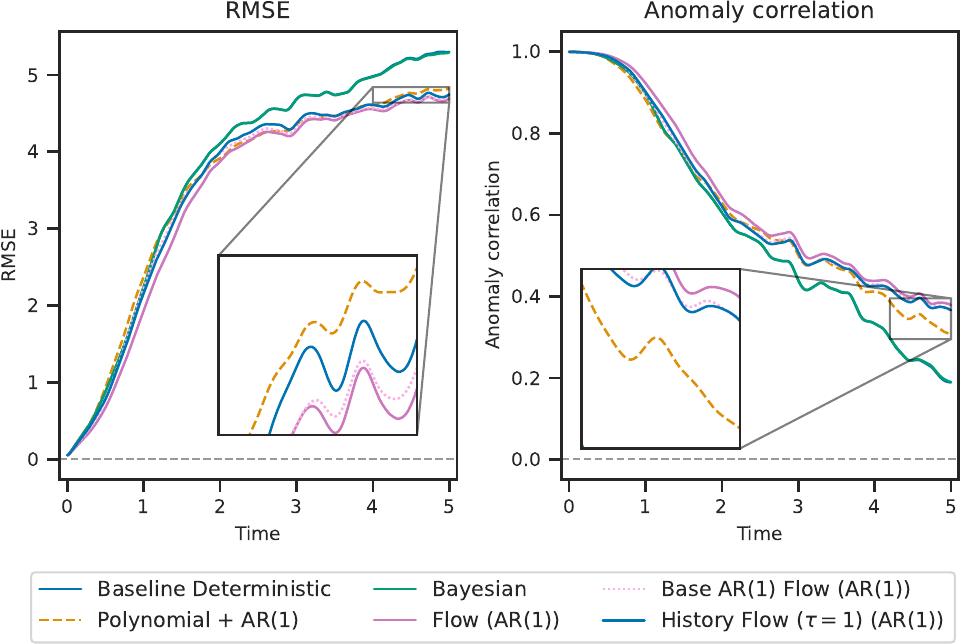}
    \caption{RMSE and anomaly correlation (lower RMSE and higher ANCR indicate better forecast skill).}
    \label{fig:rmse_ancr}
\end{figure}

\begin{figure}[t]
    \centering
    \includegraphics[width=1\linewidth]{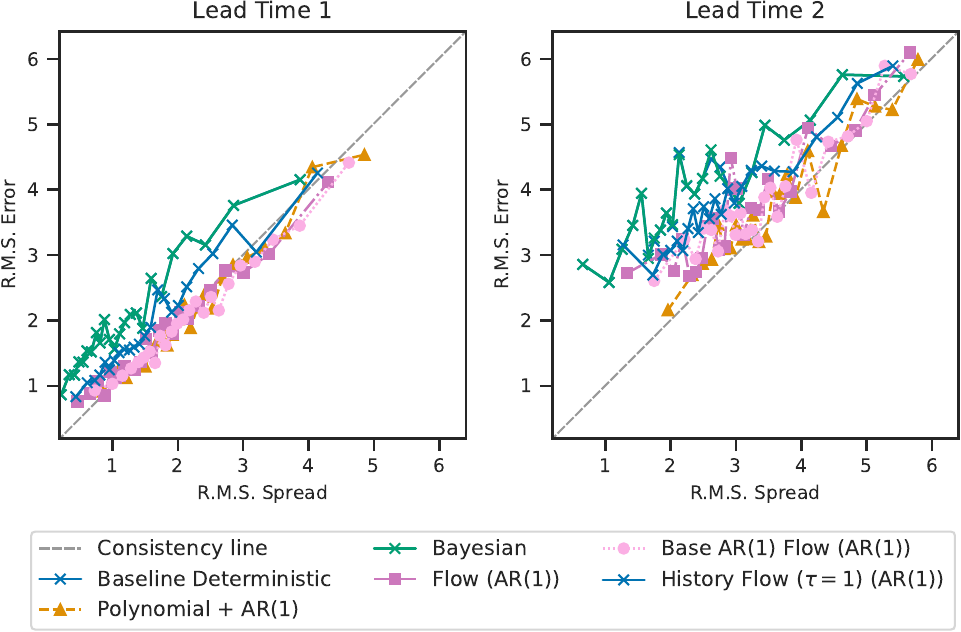}
    \caption{Spread-error consistency at lead times 1 and 2. 
    Some model results overlap in this combined view; separate model plots are shown in
    Figs.~\ref{fig:spread_error_grid_full_lt_1}--\ref{fig:spread_error_grid_full_lt_3}.
    }
    \label{fig:spread_error}
\end{figure}

We now evaluate forecast performance in the full state space 
$\mathbf{X}\in\mathbb{R}^K$.
All diagnostics are vector-based and compare the ensemble mean to the truth.
Formal definitions of root-mean-square error (RMSE), anomaly correlation (ANCR), and root-mean-square (RMS) spread are given in Appendix~\ref{sec:forecast_metrics_appendix}.

\xhdr{Forecast skill.} 
RMSE measures the magnitude of the ensemble-mean error across all components, while ANCR quantifies spatial pattern agreement. 
ANCR equals $1$ for perfect phase alignment and decreases toward $0$ as phase coherence is lost. 
RMSE, therefore, reflects error amplitude, while ANCR assesses preservation of large-scale structure. 
As shown in Fig.~\ref{fig:rmse_ancr}, the magnitude and growth of RMSE and the decay of ANCR are consistent with known L96 predictability time scales \citep{wilks2005}. 
The deterministic and Bayesian parameterizations exhibit the poorest performance, with faster error growth and more rapid loss of phase coherence. 
Stochastic models perform somewhat better, and AR(1)-sampled flow models achieve the highest overall skill (see Appendix Fig.~\ref{fig:rmse_ancr_full_appendix} for results from all models). 
In particular, the flow models with AR(1) sampling slightly outperform the polynomial+AR(1) model, suggesting that improved representation of state-dependent conditional structure helps preserve large-scale patterns. 
Although the gains are modest in this low-dimensional setting, persistent stochastic forcing appears to consistently enhance short-range forecast skill.
Differences among the flow-based models are generally small, with the normal flow with AR(1) sampling performing marginally best overall.

\xhdr{Ensemble calibration.}
Forecast reliability requires predicted probabilities to agree with observed frequencies \citep{leutbecher2008,arnold2013,mansfield2026}. 
For ensemble forecasts, this implies the ensemble spread should reflect the ensemble-mean error, a property known as statistical consistency. 
Following \citet{leutbecher2008}, we first compute the spread separately for each initial condition and lead time using the corresponding $(\nens \! \times \! \nmodel)$ ensemble.
For each lead time, the individual initial condition ensembles are sorted by spread and grouped into equally populated bins, within which we compute the RMS spread and RMSE. 
Fig.~\ref{fig:spread_error} shows the resulting spread-error relationship for lead times 1 and 2 together with the finite-ensemble consistency line, with computation details given in Appendix~\ref{sec:spread_error_appendix}.
Ensembles above this line have errors that are systematically too large relative to their spread and are therefore underdispersive, whereas ensembles below the line are overdispersive.
At both lead times 1 and 2, the deterministic, Bayesian, and history-flow models are clearly underdispersive.
Among the flow-based models, AR(1)-sampled variants are substantially closer to the consistency line than their i.i.d.-sampled counterparts, demonstrating that temporal persistence is crucial for reliable spread growth (see App. Figs.~\ref{fig:spread_error_grid_full_lt_1} and \ref{fig:spread_error_grid_full_lt_2} for all models).
The polynomial+AR(1) scheme, as well as the base AR(1) and normal flows with AR(1) sampling, are closest to the consistency line.
This suggests that calibration depends more strongly on the temporal variance growth rate than on additional conditional flexibility.
From lead time 1 to 2, all ensembles become more underdispersive, indicating that spread growth lags error growth as forecast lead time increases.
This trend continues at lead time 3 (not shown; see Appendix~\ref{fig:spread_error_grid_full_lt_3}).
At lead times 2 and 3, polynomial+AR(1) marginally outperforms the best flow models in terms of statistical consistency.
Thus, while flows improve dynamical skill as measured by RMSE and ANCR, spread-error consistency appears to be controlled primarily by explicit persistent memory in the stochastic parameterization.
We additionally verified that the separated $(\ninit \! \times \! \nens \! \times \! \nmodel)$ configuration gives the same RMSE, ANCR, and spread-error behavior as a traditional operational ensemble, where perturbations and stochastic realizations are combined (Appendix~\ref{sec:mixed_ensemble_appendix} and \ref{sec:forecast_mixed_ensemble}).
This confirms that our framework is representative of the standard ensemble setting.

\xhdr{Summary.}
Reliable ensembles require model-induced variability in addition to initial perturbations, and temporal persistence largely controls calibration by setting the spread-growth rate. Conditional flexibility yields smaller gains in dynamical skill and is secondary for calibration in L96.
gains in dynamical skill and is secondary for calibration in L96.

\section{Conclusion}

We introduced a conditional decomposition framework for diagnosing how uncertainty develops from a given initial state, complementing variance-based approaches that partition long-term ensemble spread.
Using the two-scale Lorenz '96 system, we examined how internal variability, initial-condition uncertainty, and model uncertainty jointly shape ensemble behavior.
The results show that stochastic model variability is essential for reliable forecasts: initial perturbations alone are insufficient, while temporally persistent stochasticity drives early spread growth and improves calibration.
By contrast, the additional distributional flexibility of modern generative ML methods yields only modest gains in state-dependent forecast skill, likely because L96 is a highly idealized, low-dimensional test bed with near-Gaussian statistics that favor simple closures.
Although the quantitative results may not transfer directly to operational numerical models, the framework provides a transparent way to diagnose how stochastic structure affects ensemble spread and reliability.
Overall, the results highlight the importance of temporally structured stochastic parameterizations for producing ensembles whose spread more faithfully represents forecast uncertainty.
Code for reproducing the results is publicly available at \url{https://github.com/bkueh/decomposing_uncertainty_in_lorenz96}.

\begin{disclosure}
Large Language Models (LLMs) were used to polish the text and assist with coding. 
The authors reviewed all LLM suggestions and take full responsibility for the text and code.
\end{disclosure}

\begin{contributions}
    All authors conceived the study.
    BK developed the methodology, implemented the code, performed the experiments, analyzed the results, and wrote the manuscript.
    NK and DC discussed the results and commented on the manuscript.
    All authors read and approved the final manuscript.
\end{contributions}

\begin{acknowledgements} 
    The authors would like to thank the five anonymous reviewers whose feedback helped improve the manuscript. 
    BK and NK thank Nathanael Bosch, Hannah Christensen, Maximilian Gelbrecht, Nicholas Krämer, Axel Lauer, Laura Mansfield, and Alistair White for their valuable insights and the fruitful discussions. 
    BK carried out part of this work during a PhD internship with the Scientific Computing group at CWI.
    NK has been supported by the German Federal Ministry of Education and Research (Grant: 01IS24082) and by the European Union (ERC, DYNAMICAUS, 101221985).
\end{acknowledgements}

\printbibliography

\newpage
\onecolumn

\appendix
\appendixtitle

\renewcommand{\thefigure}{A\arabic{figure}}
\setcounter{figure}{0}
\renewcommand{\thetable}{A\arabic{table}}
\setcounter{table}{0}

\section{Glossary}
\printglossary

\section{Flow Variations}
\label{sec:flow_variations_appendix}

\paragraph{Tail flow.}
As an additional variation not discussed in the main text, we consider a tail-augmented flow to improve flexibility in the
representation of extreme events. 
Normalizing flows with Gaussian base distributions can struggle to
represent heavy-tailed behavior \citep{jaini2020}. 
To increase flexibility in the tails without altering the core
architecture, we append a monotone tail transformation to the
RealNVP output, following \cite{hickling2025}.

Let $\mathbf{Y}_t$ denote the output of the RealNVP core. 
The final closure is defined as
\begin{align*}
\mathbf{U}_t = R_\phi(\mathbf{Y}_t),
\end{align*}
where $R_\phi$ acts element-wise. For component $d$,
\begin{align*}
R(y_d)
=
\mu_d
+
\sigma_d
\frac{s_d}{\lambda_{s}}
\left[
\operatorname{erfc}\!\left(\frac{|y_d|}{\sqrt{2}}\right)^{-\lambda_{s}}
-
1
\right],
\qquad
s_d = \operatorname{sign}(y_d),
\end{align*}
with \(\lambda_s=\lambda^+\) for \(s_d>0\) and \(\lambda^-\) otherwise.
The parameters satisfy \(\sigma_d>0\), \(\lambda^+>0\), and
\(\lambda^->0\), and are learned jointly with the flow parameters.
We initialize the shared tail parameters as
$\lambda^+ = \lambda^- = 0.1$.

The transformation is monotone with a closed-form inverse and tractable
Jacobian determinant, so likelihood evaluation remains exact.
It effectively reshapes Gaussian tails into heavier tails with
tunable asymmetry, while avoiding heavy-tailed inputs to the neural
network layers.
Results for the tail flow are reported in the appendix figures
together with the other flow variants.

\paragraph{AR(1) base flow.}
In Sec.~\ref{sec:parameterizations}, we briefly introduced a flow variant in which the base distribution itself is defined as an AR(1) process in latent space.
Here we provide additional details.

Instead of assuming an i.i.d.\ Gaussian base, we impose for each latent dimension
\begin{align*}
\mathbf{Z}_t
=
\rho\,\mathbf{Z}_{t-1}
+
\sigma\,\boldsymbol{\varepsilon}_t,
\qquad
\boldsymbol{\varepsilon}_t \sim \mathcal{N}(\mathbf{0},\mathbf{I}),
\end{align*}
with $|\rho|<1$ and $\sigma>0$.
The initial state is taken as
$\mathbf{Z}_1 \sim \mathcal{N}(\mathbf{0},\mathbf{I})$.
The parameters $(\rho,\sigma)$ are learned jointly with the flow
parameters.

Given a training sequence
$\{(\mathbf{X}_t,\mathbf{U}_t)\}_{t=1}^T$, we invert the conditional
flow pointwise in time,
\begin{align*}
\mathbf{Z}_t = \mathbf{g}_\theta^{-1}(\mathbf{U}_t,\mathbf{c}_t),
\end{align*}
and maximize the corresponding joint sequence log likelihood
\begin{align*}
\log p_{\theta,\rho,\sigma}(\mathbf{U}_{1:T}\mid \mathbf{c}_{1:T})
=
\log p_{\rho,\sigma}(\mathbf{Z}_{1:T})
+
\sum_{t=1}^T
\log \left|
\det
\frac{\partial \mathbf{g}_\theta^{-1}}{\partial \mathbf{U}_t}
\right|,
\end{align*}
where $p_{\rho,\sigma}(\mathbf{Z}_{1:T})$ denotes the Gaussian AR(1)
sequence density. 
The flow transformation is evaluated separately at each time step, so the change-of-variables term decomposes into the sum of single-step Jacobian contributions above. 
The latent base density, however, is evaluated as a sequence density rather than as a product of independent standard-normal terms. 
In our implementation, the initial latent state is scored under the standard-normal prior, while subsequent states are scored through their AR(1) innovations:
\begin{align*}
\log p_{\rho,\sigma}(\mathbf{Z}_{1:T})
=
\log p(\mathbf{Z}_1)
+
\sum_{t=2}^{T}
\log p_{\rho,\sigma}(\mathbf{Z}_t \mid \mathbf{Z}_{t-1}),
\end{align*}
with
\begin{align*}
\mathbf{Z}_1 &\sim \mathcal{N}(\mathbf{0},\mathbf{I}),\\
\mathbf{Z}_t \mid \mathbf{Z}_{t-1}
&\sim
\mathcal{N}(\rho \mathbf{Z}_{t-1}, \sigma^2 \mathbf{I}).
\end{align*}
Thus, temporal dependence enters through the latent sequence likelihood, while the conditional flow itself remains pointwise in time.

During simulation, we evolve the latent state externally according to the learned AR(1) dynamics using the same parameters $(\rho,\sigma)$ as in the base density.
At each sampling interval, the current latent state $\mathbf{Z}_t$ is passed to the conditional flow together with the current conditioning state, and the subgrid tendency is sampled as
$\mathbf{U}_t=\mathbf{g}_\theta(\mathbf{Z}_t,\mathbf{c}_t)$.
Thus, although the flow is evaluated one time step at a time, the latent samples retain the temporal dependence implied by the learned AR(1) process.
Using the same AR parameters in the sequence likelihood and during rollout aligns the temporal structure used for training and sampling.

\section{Parameterization Training and Flow Configuration}
\label{sec:flow_configuration_appendix}

All parameterization schemes are trained on the same dataset with identical sample sizes. 
The data are generated from a long integration of the fully resolved L96 system over 20{,}000 MTU: the first 3{,}000 are used for training, the next 2{,}000 for validation, and the remaining 15{,}000 for testing. 
Only the flow models use the validation data during training.  

The flow models all follow the conditional RealNVP architecture \citep{dinh2017} with four affine coupling layers. 
Each coupling layer uses alternating binary masks and is conditioned on an input variable $\mathbf{c}$ \citep{papamakarios2017, trippe2018}. 
The scale and translation functions are parameterized by identical multilayer perceptrons (MLPs) with three hidden layers of 128 units each and ReLU activations.

Each coupling layer transforms $\mathbf{x} \in \mathbb{R}^d$ as
\begin{align*}
\mathbf{y} = \mathbf{x}_{\text{mask}} + (1 - \mathbf{m}) \odot \left( \mathbf{x} \odot \exp\big(s(\mathbf{x}_{\text{mask}}, \mathbf{c})\big) + t(\mathbf{x}_{\text{mask}}, \mathbf{c}) \right),
\end{align*}
where $\mathbf{m} \in \{0,1\}^d$ is a binary mask, $\mathbf{x}_{\text{mask}} = \mathbf{x} \odot \mathbf{m}$ denotes the masked input, $\odot$ denotes element-wise multiplication, and $s(\cdot)$ and $t(\cdot)$ are the outputs of the MLPs.

We train the models using the Adam optimizer \citep{kingma2017} with a learning rate of $10^{-3}$, a batch size of 2056, and a maximum of 100 epochs. 
Early stopping is applied with a patience of 8 epochs based on the validation loss.

\section{Comparison with Mixed Ensemble Configuration}
\label{sec:mixed_ensemble_appendix}

\begin{figure}
    \centering
    \includegraphics[width=0.8\linewidth]{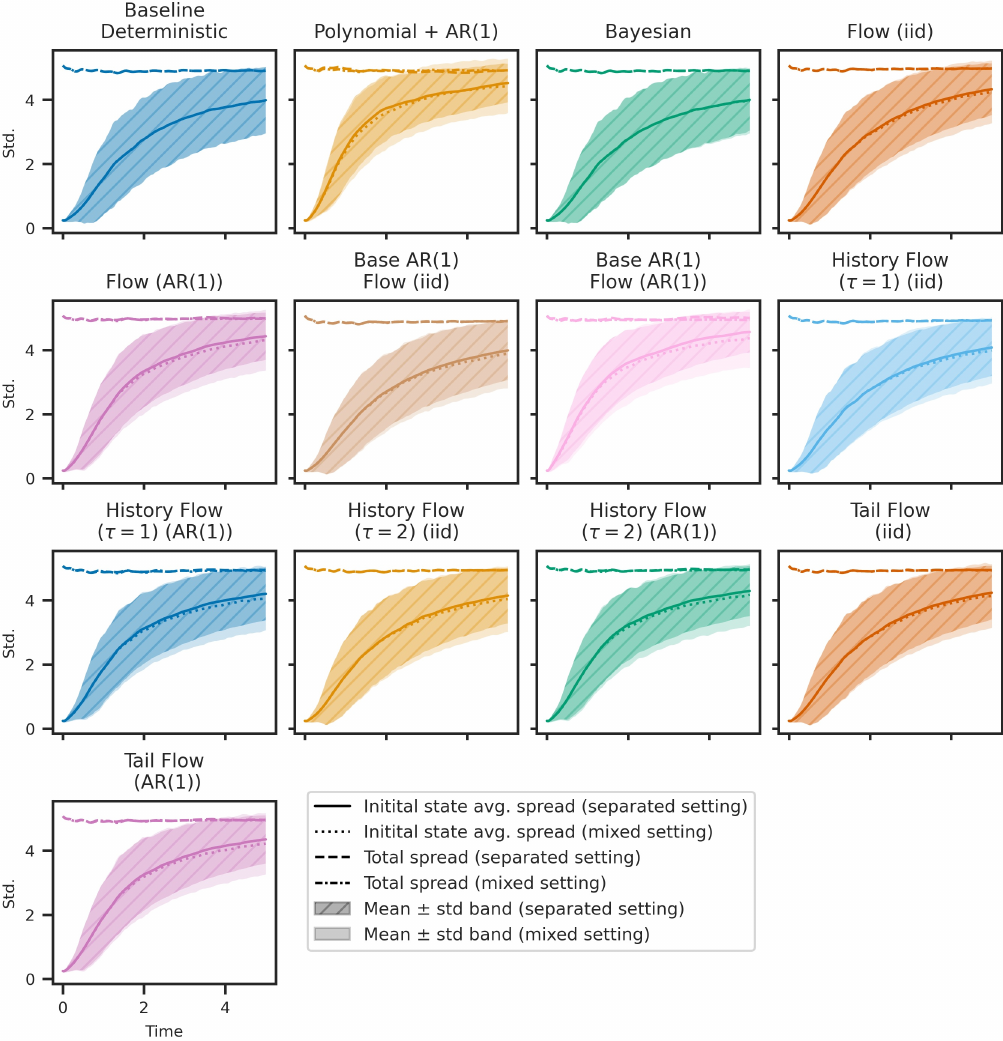}
    \caption{Comparison of total initial-state-averaged ensemble spread in the separated configuration $(\ninit \! \times \! \nens \! \times \! \nmodel)$ and the traditional mixed configuration $(\ninit \! \times \! \nens)$ with stochastic physics active, averaged over $k$.
    The ensemble mean and total spread are nearly identical in both settings, since they sample the same underlying sources of variability.
    However, spread magnitude decreases in the $(\ninit \! \times \! \nens \! \times \! \nmodel)$ configuration because explicit sampling over stochastic realizations reduces sampling variability and yields a more stable estimate of total spread.}
    \label{fig:spread_mix_vs_full_avg_k_appendix}
\end{figure}

\begin{figure}
    \centering
    \includegraphics[width=0.8\linewidth]{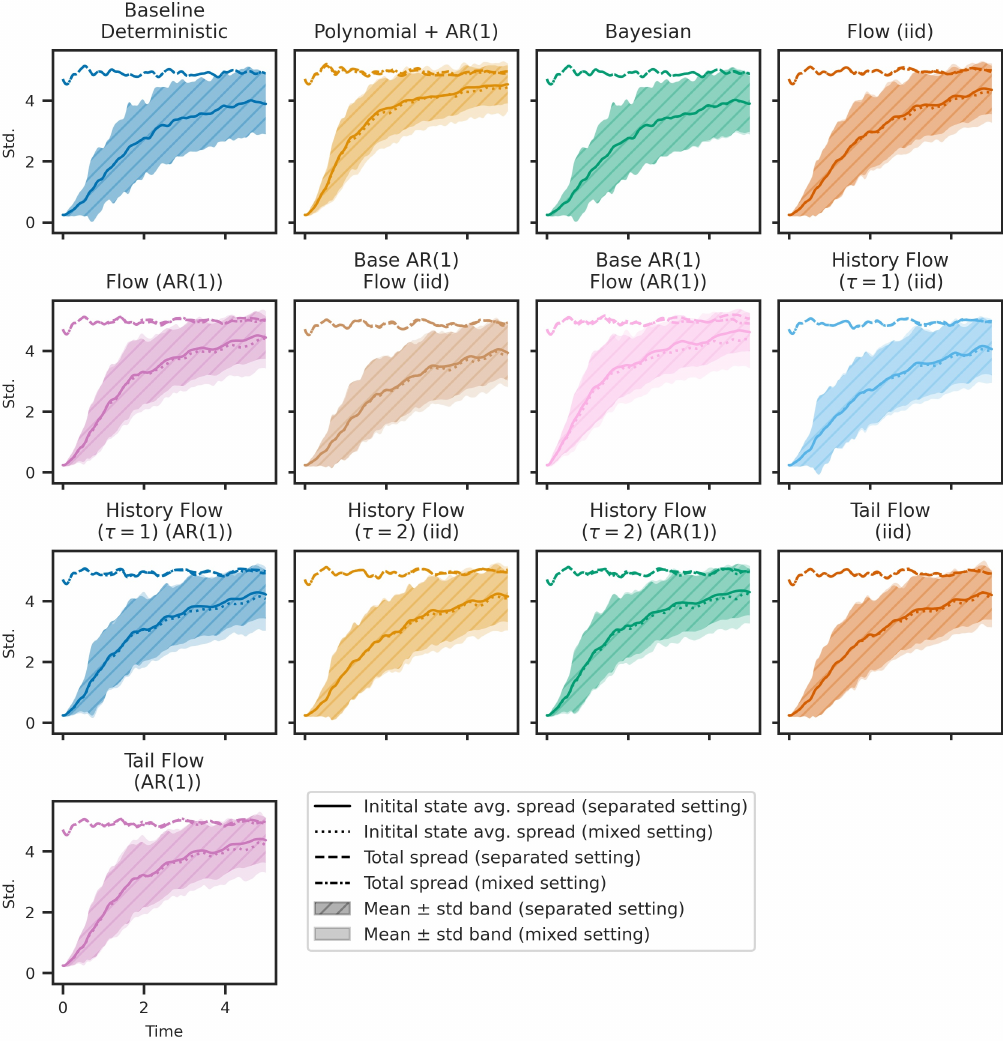}
    \caption{Same comparison as in Fig.~\ref{fig:spread_mix_vs_full_avg_k_appendix}, but for the single spatial index $k=0$ rather than averaged over $k$.
    Single-$k$ results are noisier but show the same qualitative behavior as the spatially averaged results.
    Other spatial indices exhibit similar behavior.}
    \label{fig:spread_mix_vs_full_single_k_appendix}
\end{figure}

In a traditional mixed ensemble $(\ninit \times \nens)$ with stochastic physics active, initial-condition perturbations and stochastic model realizations are combined within a single ensemble index. 
Each perturbed initial state is therefore integrated with a different stochastic realization. 
As a consequence, the contributions from initial perturbations and stochastic variability cannot be disentangled.
In this setting, only the total conditional spread
\begin{align}
V^{\mathrm{total,mixed},i}_{k}(t)
=
\mathbb{E}_{i}
\left[
\operatorname{Var}_{j}
\!\left(X_{i,j,k}(t)\right)
\right]
\end{align}
is directly observable, and the individual contributions from initial perturbations and stochastic variability cannot be separated. 
The corresponding ensemble mean is
\begin{align}
\mu_{k} = \mathbb{E}_{i,j} \left[ X_{i,j,k}\right].
\end{align}

To ensure that the results of our spread decomposition are representative of operational ensemble configurations, we compare this mixed configuration with the full ensemble setting considered in the main paper, in which stochastic realizations are sampled explicitly, enabling a formal attribution of uncertainty. 
In the separated case, the total spread (see Sec.~\ref{sec:uncertainty_decomposition}) is given by
\begin{align}
    V^{\mathrm{total},i}_{k}(t)
    =
    \mathbb{E}_{i}
    \left[
    \operatorname{Var}_{j,m}
    \!\left(X_{i,j,m,k}(t)\right)
    \right]
\end{align}
and the ensemble mean becomes
\begin{align}
\mu_{k} = \mathbb{E}_{i,j,m} \left[ X_{i,j,m,k}\right].
\end{align}
The results are shown in Fig.~\ref{fig:spread_mix_vs_full_avg_k_appendix} for averages over $k$ and in Fig.~\ref{fig:spread_mix_vs_full_single_k_appendix} for the single spatial index $k=0$. 
In both configurations, the ensemble mean and total spread are nearly identical, reflecting that both sample the same underlying sources of variability. 
Small differences arise in the $(N_{\mathrm{init}} \times N_{\mathrm{ens}} \times N_{\mathrm{model}})$ configuration, where the spread magnitude slightly contracts starting at approximately $t \approx 2$. 
This is due to the explicit sampling over stochastic realizations $m$, which reduces sampling variability in the estimate of total spread.
Overall, these results indicate that the mixed and full ensemble configurations are statistically comparable with respect to mean and total variability.

\section{Distributional Metrics}
\label{sec:distributional_metric_appendix}

For sufficiently long integrations, trajectories sample the invariant
measure of the system.
We therefore aggregate all spatial components, time steps, and trajectories into a single one-dimensional sample for both truth and model.

Using shared histogram bins, we estimate marginal densities
$\hat p(x)$ (truth) and $\hat q(x)$ (model).
The Hellinger distance is computed as
\begin{align}
H(\hat p,\hat q)
=
\sqrt{
\frac{1}{2}
\sum_b
\left(
\sqrt{\hat p_b}
-
\sqrt{\hat q_b}
\right)^2
\Delta x_b
},
\end{align}
where $b$ indexes histogram bins and $\Delta x_b$ denotes bin width.
This metric is symmetric and bounded between $0$ and $1$.

As a complementary statistic, we compute the two-sample
Kolmogorov-Smirnov (KS) statistic
\begin{align}
D_{\mathrm{KS}}
=
\sup_x
\left|
F_{\mathrm{truth}}(x)
-
F_{\mathrm{model}}(x)
\right|,
\end{align}
where $F$ denotes the empirical cumulative distribution function.
The KS statistic measures the maximum deviation between
cumulative distributions and is particularly sensitive to
systematic biases in location and spread.

Both the Hellinger distances and KS statistics for all models are shown in Tab.~\ref{tab:hellinger_ks_appendix}.

\begin{table}[t]
\centering
\caption{Distributional distances to truth for all reduced models.}
\label{tab:hellinger_ks_appendix}
\begin{tabular}{lcc}
\toprule
Model & Hellinger Distance & Kolmogorov-Smirnov Statistic \\
\midrule
Baseline Deterministic & 0.0416 & 0.0160 \\
Polynomial + AR(1) & 0.0311 & 0.0124 \\
Bayesian & 0.0417 & 0.0160 \\
Flow (iid) & 0.0676 & 0.0210 \\
Flow (AR(1)) & 0.0570 & 0.0194 \\
History Flow ($\tau = 1$) (iid) & 0.0808 & 0.0252 \\
History Flow ($\tau = 1$) (AR(1)) & 0.0715 & 0.0242 \\
History Flow ($\tau = 2$) (iid) & 0.0649 & 0.0209 \\
History Flow ($\tau = 2$) (AR(1)) & 0.0567 & 0.0197 \\
Base AR(1) Flow (iid) & 0.0915 & 0.0273 \\
Base AR(1) Flow (AR(1)) & 0.0517 & 0.0191 \\
Tail Flow (iid) & 0.0701 & 0.0223 \\
Tail Flow (AR(1)) & 0.0599 & 0.0207 \\
\bottomrule
\end{tabular}
\end{table}

\section{Sensitivity Metrics}
\label{sec:sensitivity_appendix}
To quantify perturbation growth within each initial state ensemble of the fully resolved two-scale L96 system, we compute the average pairwise Euclidean distance (APD) between ensemble members,
\begin{align}
\mathrm{APD}(t)
=
\frac{1}{\ninit}
\sum_i
\frac{2}{\nens(\nens-1)}
\sum_{j<\ell}
\|\mathbf{X}^{\mathrm{true}}_{i,j}(t)-\mathbf{X}^{\mathrm{true}}_{i,\ell}(t)\|_2, 
\end{align}
where $\|\cdot\|_2$ denotes the L2 norm. 
Here $i=1,\ldots,\ninit$ indexes distinct perfect initial states, and
$j,\ell \in \{1,\ldots,\nens\}$ with $j<\ell$ index distinct ensemble members generated from the same initial state.
APD isolates divergence due to initial-condition perturbations
by averaging distances within each $i$ and subsequently across
initial states. 
It provides a direct diagnostic of short-term
chaotic separation independent of mean shifts across the attractor.

\section{Forecast Metrics}
\label{sec:forecast_metrics_appendix}

\begin{figure}
    \centering
    \includegraphics[width=0.8\linewidth]{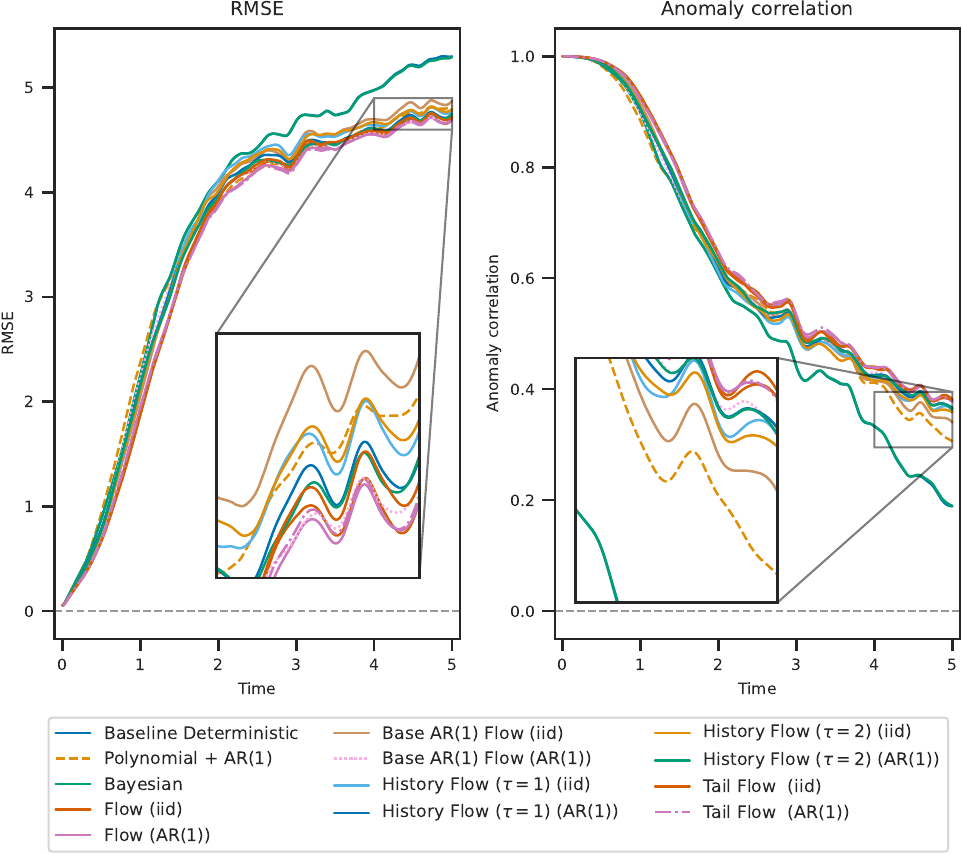}
    \caption{RMSE and anomaly correlation (ANCR) in state space for the separated ensemble configuration $(\ninit \! \times \! \nens \! \times \! \nmodel)$.
    This figure extends Fig.~\ref{fig:rmse_ancr} by including additional flow variants and their i.i.d.\ sampling counterparts.
    Lower RMSE and higher ANCR indicate better forecast skill.
    The i.i.d.-sampled flows perform slightly worse than their AR(1)-sampled counterparts, but the clearest separation is between the reduced models using the deterministic and Bayesian parameterizations, which show the lowest forecast skill, and the remaining stochastic reduced models.}
    \label{fig:rmse_ancr_full_appendix}
\end{figure}

\begin{figure}
    \centering
    \includegraphics[width=0.8\linewidth]{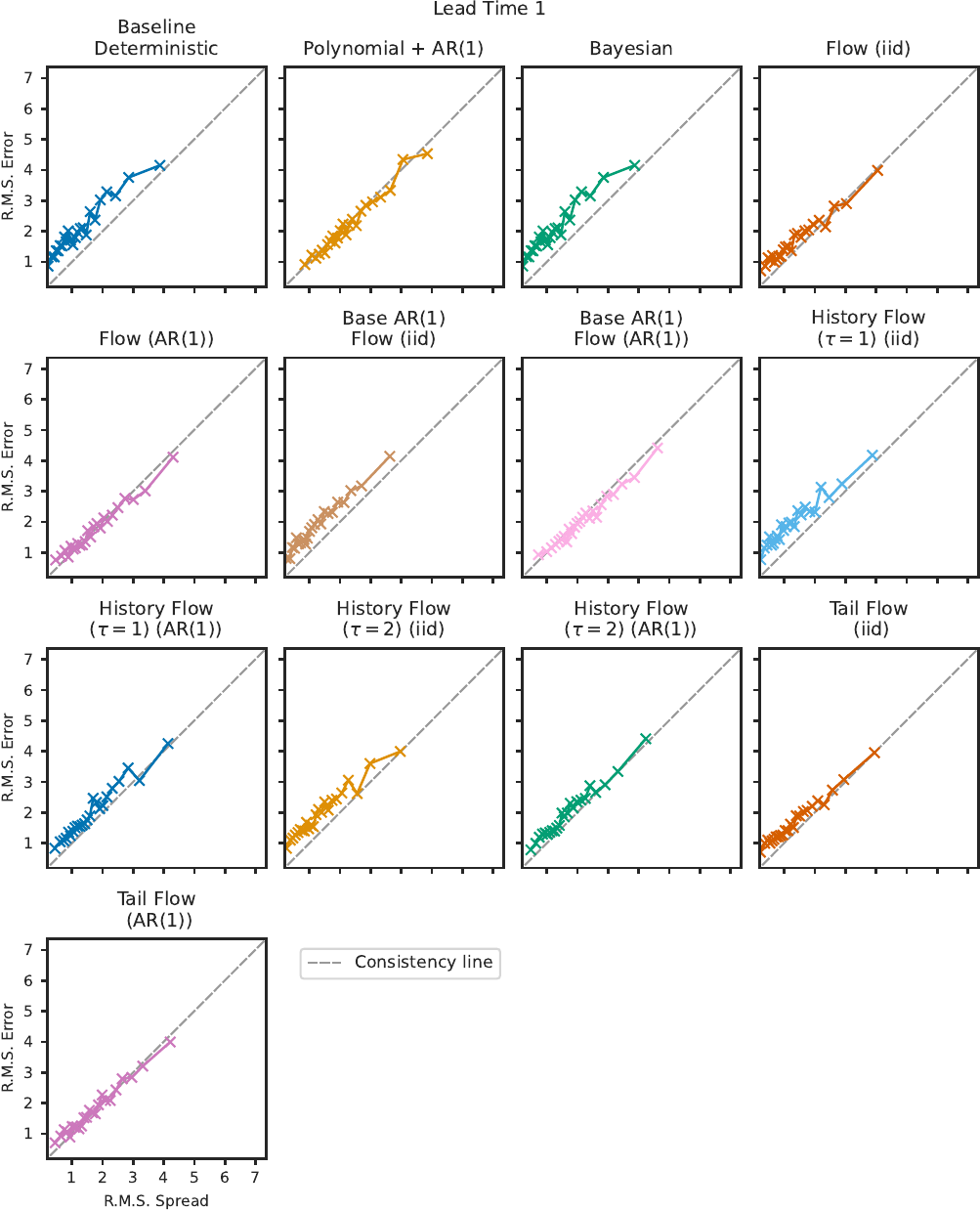}
    \caption{Spread-error consistency at lead time $\tau=1$ for the separated ensemble configuration $(\ninit \! \times \! \nens \! \times \! \nmodel)$. 
    This extends Fig.~\ref{fig:spread_error} by including additional flow variants and their i.i.d.-sampling counterparts. 
    Diagnostics are computed using 24 bins with 100 samples per bin. 
    Models above the consistency line are underdispersive, whereas models below it are overdispersive. 
    The deterministic and Bayesian models are underdispersive and nearly identical. 
    Among the flow models, AR(1) sampling improves spread-error consistency relative to i.i.d. sampling. 
    The best-performing models are polynomial+AR(1), normal flow, base AR(1) flow, and tail flow, all with AR(1) sampling.
    }
    \label{fig:spread_error_grid_full_lt_1}
\end{figure}

\begin{figure}
    \centering
    \includegraphics[width=0.8\linewidth]{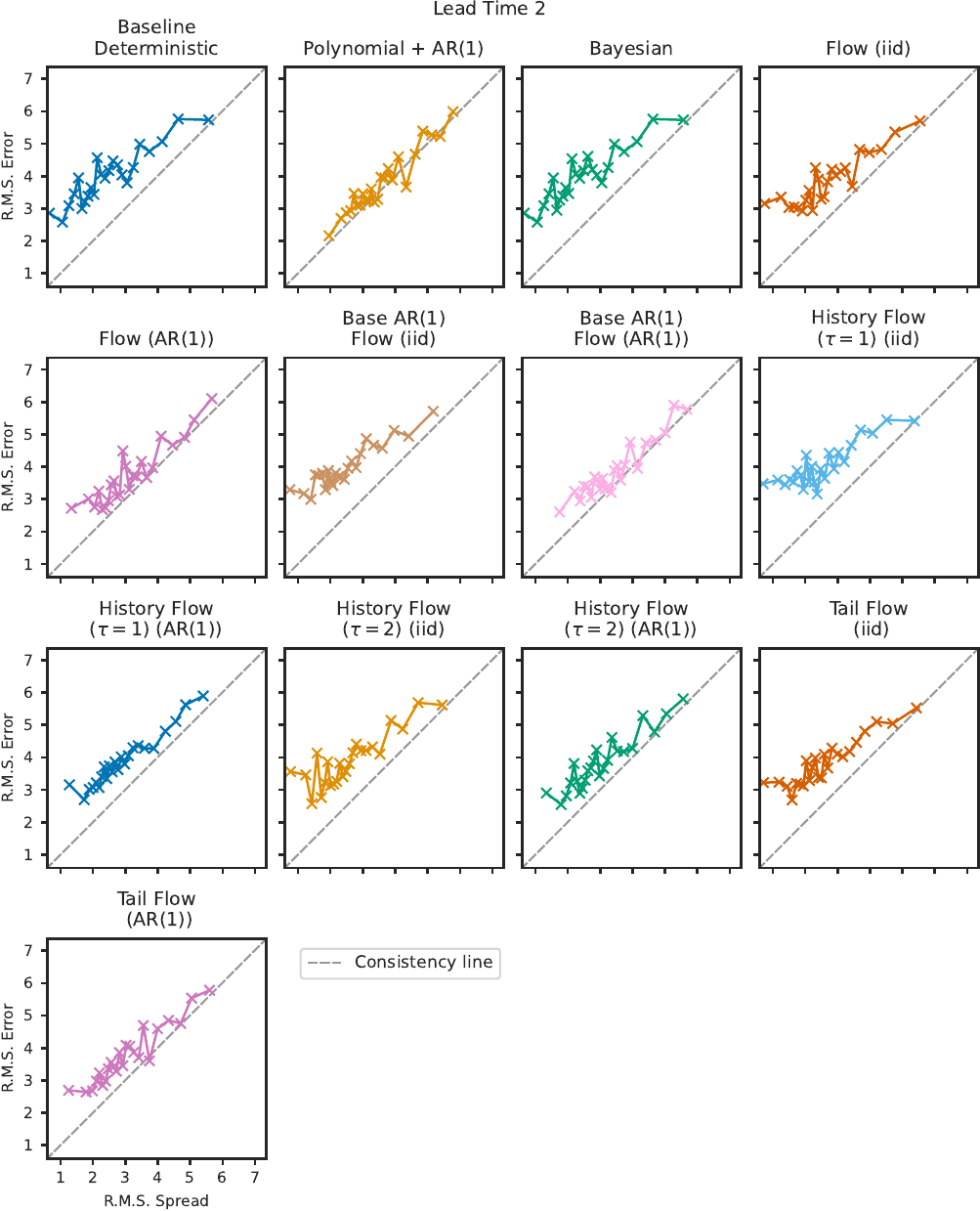}
    \caption{Spread-error consistency at lead time $\tau=2$ for the separated ensemble configuration $(\ninit \! \times \! \nens \! \times \! \nmodel)$. 
    Compared to $\tau=1$ (Fig.~\ref{fig:spread_error_grid_full_lt_1}), all models move farther from the consistency line and become more underdispersive. 
    The polynomial+AR(1) model shows the closest agreement, while the normal flow, base AR(1) flow, and tail flow with AR(1) sampling also remain relatively close to the consistency line.
    }
    \label{fig:spread_error_grid_full_lt_2}
\end{figure}

\begin{figure}
    \centering
    \includegraphics[width=0.8\linewidth]{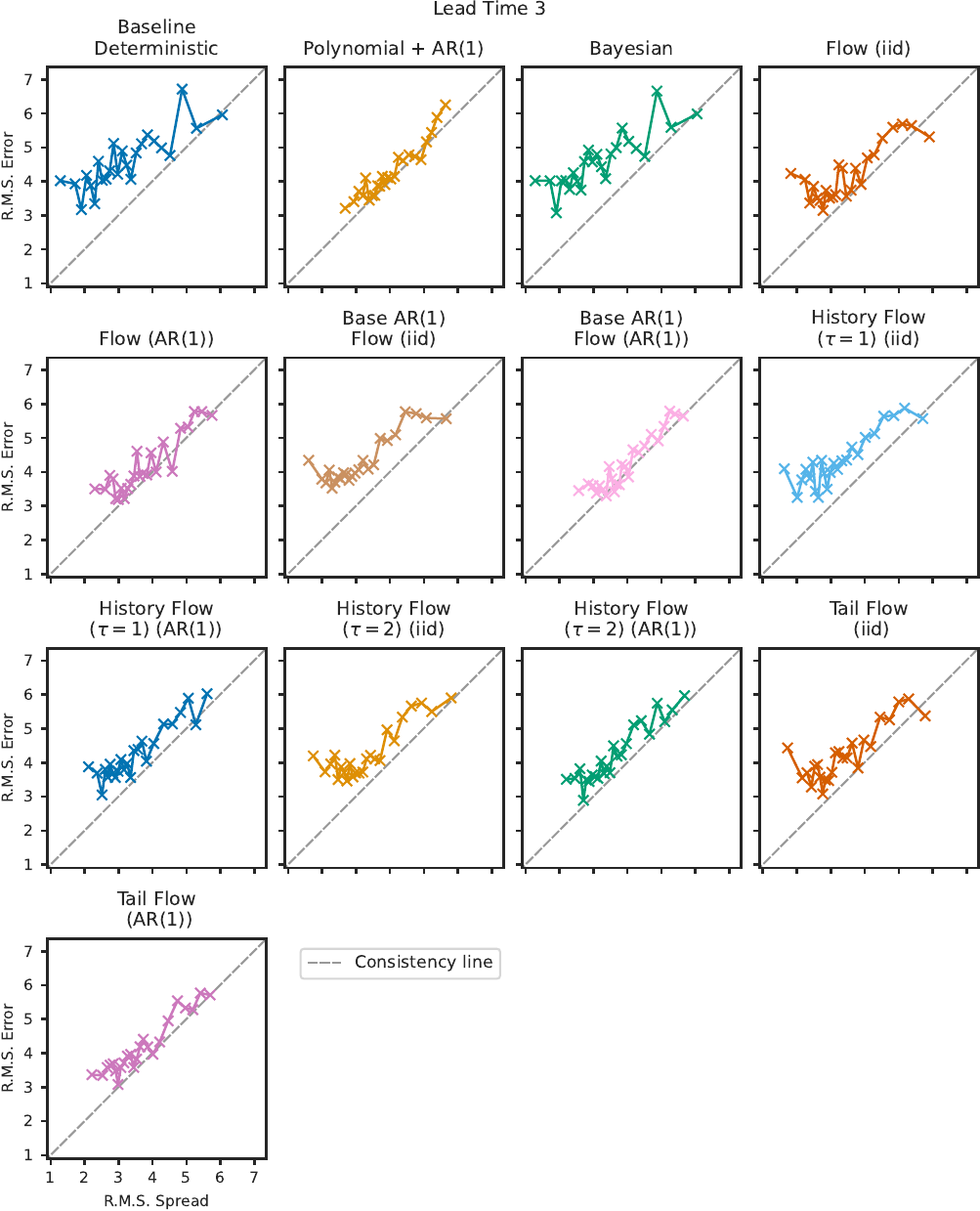}
    \caption{Spread--error consistency at lead time $\tau=3$ for the separated ensemble configuration $(\ninit \! \times \! \nens \! \times \! \nmodel)$. 
    Compared with lead times $\tau=1$ and $\tau=2$ (Figs.~\ref{fig:spread_error_grid_full_lt_1} and \ref{fig:spread_error_grid_full_lt_2}), only polynomial+AR(1) and the base AR(1) flow remain approximately consistent. 
    Across models, bin values shift from the lower-left to the upper-right, indicating increasing ensemble error.
    }
    \label{fig:spread_error_grid_full_lt_3}
\end{figure}

\begin{table}[t]
\centering
\caption{RMSE and ANCR at lead time $t=5$ for the separated ensemble configuration $(\ninit \! \times \! \nens \! \times \! \nmodel)$ and the mixed configuration $(\ninit \! \times \! \nens)$. 
The three best values in each column are shown in bold. 
Since these values represent only a single lead time, they do not capture the full temporal evolution of RMSE and ANCR and should be read as a numerical supplement to Figs.~\ref{fig:rmse_ancr_full_appendix} and \ref{fig:rmse_ancr_mix_appendix}.}

\label{tab:full_vs_mix_rmse_ancr_appendix}
\begin{tabular}{lcccc}
\toprule
& \multicolumn{2}{c}{Separated configuration} & \multicolumn{2}{c}{Mixed configuration} \\[0.2cm]
Model & RMSE & ANCR & RMSE & ANCR \\
\midrule
Baseline Deterministic             & 5.2997 & 0.1888 & 5.2997 & 0.1888 \\
Polynomial + AR(1)                 & 4.8266 & 0.3062 & 4.8858 & 0.2818 \\
Bayesian                           & 5.2927 & 0.1898 & 5.3109 & 0.1862 \\
Flow (iid)                         & 4.6931 & \textbf{0.3812} & \textbf{4.7823} & \textbf{0.3625} \\
Flow (AR(1))                       & \textbf{4.6921} & \textbf{0.3788} & \textbf{4.7725} & \textbf{0.3600} \\
History Flow ($\tau =1$) (iid)     & 4.7799 & 0.3631 & 4.8525 & 0.3477 \\
History Flow ($\tau =1$) (AR(1))   & 4.7472 & 0.3660 & 4.8164 & 0.3591 \\
History Flow ($\tau =2$) (iid)     & 4.7950 & 0.3584 & 4.8989 & 0.3399 \\
History Flow ($\tau =2$) (AR(1))   & 4.7501 & 0.3635 & 4.8455 & 0.3399 \\
Base AR(1) Flow (iid)              & 4.8720 & 0.3400 & 4.9513 & 0.3328 \\
Base AR(1) Flow (AR(1))            & \textbf{4.7153} & 0.3660 & 4.8219 & 0.3370 \\
Tail Flow (iid)                    & 4.7182 & \textbf{0.3783} & \textbf{4.7543} & \textbf{0.3730} \\
Tail Flow (AR(1)                   & \textbf{4.7020} & 0.3773 & 4.8057 & 0.3503 \\
\bottomrule
\end{tabular}
\end{table}

To assess forecast quality, we compute forecast metrics for the separated ensemble configuration $(\ninit \! \times \! \nens \! \times \! \nmodel)$ and truth $(\ninit)$, i.e., a single fully resolved integration from each perfect initial state. 

\paragraph{Root mean square error (RMSE).}
Forecast accuracy is quantified using the state-space RMSE of the ensemble mean,
\begin{align}
\mathrm{RMSE}(t)
=
\sqrt{
\frac{1}{\ninit \cdot K}
\sum_{i,k}
\left(
X^{\mathrm{mean}_{j,m}}_{i,k}(t)
-
X^{\mathrm{true}}_{i,k}(t)
\right)^2
}, \label{eq:rmse_appendix}
\end{align}
where $X^{\mathrm{mean}_{j,m}}$ denotes averaging over ensemble
members $j$ and model realizations $m$.
RMSE measures the magnitude of the ensemble-mean error across all spatial components.
Results for all models are shown in Fig.~\ref{fig:rmse_ancr_full_appendix}.

\paragraph{Anomaly correlation (ANCR).}
Following \citet{crommelin2008}, we additionally assess spatial pattern agreement using the anomaly correlation.
Anomalies are defined relative to the long-term truth mean
$\langle \mathbf{X}^{\mathrm{true}} \rangle$:
\begin{align}
\mathbf{a}^{\mathrm{true}}_i(t)
&=
\mathbf{X}^{\mathrm{true}}_i(t)
-
\langle \mathbf{X}^{\mathrm{true}} \rangle, \\
\mathbf{a}^{\mathrm{mean}}_i(t)
&=
\mathbf{X}^{\mathrm{mean}_{j,m}}_i(t)
-
\langle \mathbf{X}^{\mathrm{true}} \rangle, \label{eq:a_mean_appendix}
\end{align}
where $\langle \mathbf{X}^{\mathrm{true}} \rangle$ denotes the time- and initial-state mean of the long truth integration.
The anomaly correlation at lead time $t$ is then defined as 
\begin{align}
\mathrm{ANCR}(t)
=
\frac{1}{\ninit}
\sum_{i}
\frac{
\mathbf{a}^{\mathrm{true}}_i(t)
\cdot
\mathbf{a}^{\mathrm{mean}}_i(t)
}{\sqrt{
\left|
\mathbf{a}^{\mathrm{true}}_i(t)
\right|^2
\left|
\mathbf{a}^{\mathrm{mean}}_i(t)
\right|^2}
}. 
\end{align}
Here $\mathbf{a} \cdot \mathbf{b} = \sum_k a_k b_k$ and
$|\mathbf{a}|^2 = \mathbf{a} \cdot \mathbf{a}$.
ANCR measures alignment of anomaly patterns in the full state space
It equals $1$ for perfect spatial agreement, decreases as trajectories diverge, and approaches $0$ when no linear relationship remains.
Thus, while RMSE primarily reflects amplitude error, ANCR captures loss of large-scale pattern coherence. 
Results for all models are shown in Fig.~\ref{fig:rmse_ancr_full_appendix}. 

\paragraph{Spread-error relationship.}
\label{sec:spread_error_appendix}

To assess ensemble calibration, we follow the spread-error framework of \citet{leutbecher2008}. 
For each initial condition $i$, lead time $\tau$, and state component $k$, we first compute the ensemble spread as the standard deviation across ensemble members and model realizations,
\begin{align}
s_{i,k}(\tau)
= 
\sqrt{ 
\frac{1}{\nens \cdot \nmodel} \sum_{j,m} 
\left( 
X_{i,j, m, k}(\tau) - X^{\mathrm{mean}_{j,m}}_{i,k}(\tau) 
\right)^2
}, \label{eq:spread_error_spread_appendix}
\end{align}
and the corresponding ensemble-mean error,
\begin{align}
e_{i,k}(\tau)
=
X^{\mathrm{mean}_{j,m}}_{i,k}(\tau) - X^{\mathrm{true}}_{i,k}(\tau),
\label{eq:spread_error_diff_appendix}
\end{align}
where $X^{\mathrm{mean}_{j,m}}$ denotes the average over ensemble members and model realizations. 
For a fixed lead time, the samples $\{s_{i,k}(\tau)\}$ from all initial conditions and state components are sorted by spread and partitioned into equally populated bins of size 100. 
Let $\mathcal{B}_b$ denote the set of samples belonging to bin $b$.
Following the computation in \citep{mansfield2026}, the RMS spread for bin $b$ is then given by 
\begin{align}
\mathrm{RMS\,Spread}_b(\tau)
=
\frac{1}{|\mathcal{B}_b|}
\sum_{(i,k)\in\mathcal{B}_b}
s_{i,k}(\tau),
\end{align}
The RMSE for bin $b$ is
\begin{align}
\mathrm{RMSE}_b(\tau)
=
\sqrt{
\frac{1}{|\mathcal{B}_b|}
\sum_{(i,k)\in\mathcal{B}_b}
e_{i,k}(\tau)^2
}.
\end{align}

For a statistically consistent ensemble, RMS spread and RMSE satisfy the finite-ensemble relation \citep{leutbecher2008}
\begin{align}
\mathrm{RMS\,Spread}
=
\sqrt{\frac{M-1}{M+1}}
\,\mathrm{RMSE},
\end{align}
where $M$ is the ensemble size.
In the $(\ninit \! \times \! \nens \! \times \! \nmodel)$ ensemble configuration, $M = \nens \cdot \nmodel$.  
The spread-error relationships for lead times 1, 2, and 3 for all models in the separated ensemble configuration are shown in Figs.~\ref{fig:spread_error_grid_full_lt_1}, \ref{fig:spread_error_grid_full_lt_2}, and \ref{fig:spread_error_grid_full_lt_3}, respectively.  

\paragraph{Mixed ensemble configuration.}
\label{sec:forecast_mixed_ensemble}

\begin{figure}[ht]
    \centering
    \includegraphics[width=0.8\linewidth]{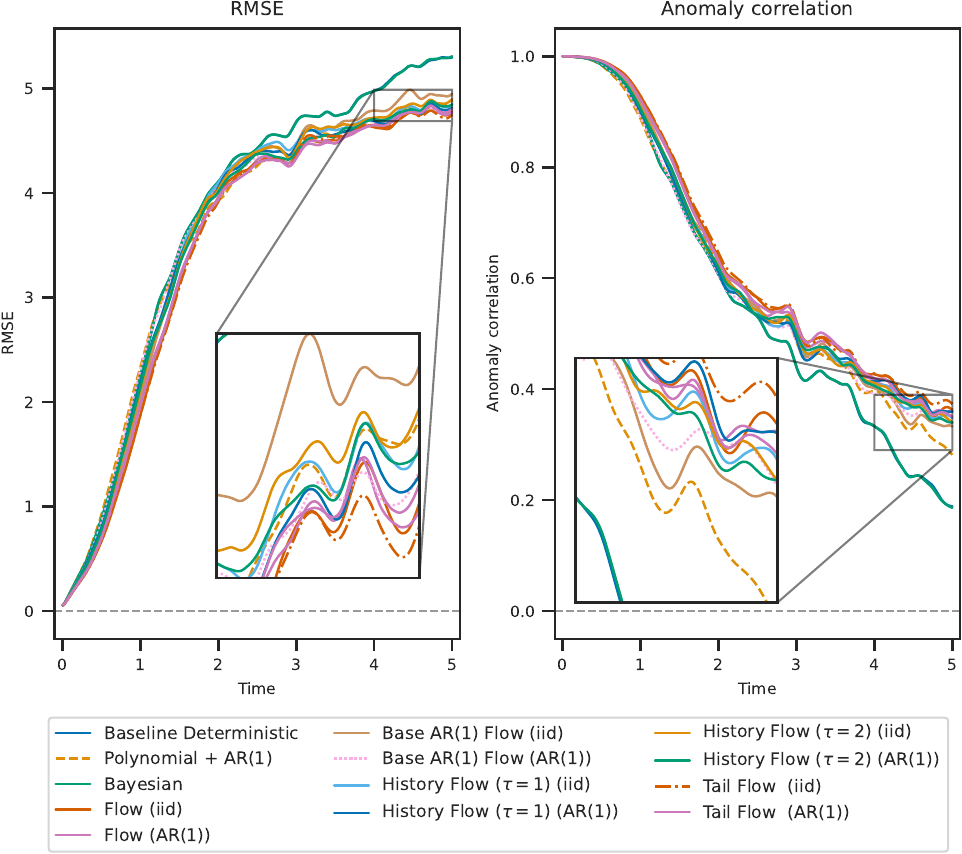}
    \caption{RMSE and anomaly correlation (ANCR) in state space for the traditional mixed configuration $(\ninit \! \times \! \nens)$ with stochastic physics active.
    This figure provides the mixed-ensemble counterpart to Fig.~\ref{fig:rmse_ancr_full_appendix}.
    Lower RMSE and higher ANCR indicate better forecast skill.
    The results are consistent with the separated ensemble configuration, showing that the separated framework preserves the forecast-skill behavior of the operationally used mixed setting.}
    \label{fig:rmse_ancr_mix_appendix}
\end{figure}

\begin{figure}[ht]
    \centering
    \includegraphics[width=0.8\linewidth]{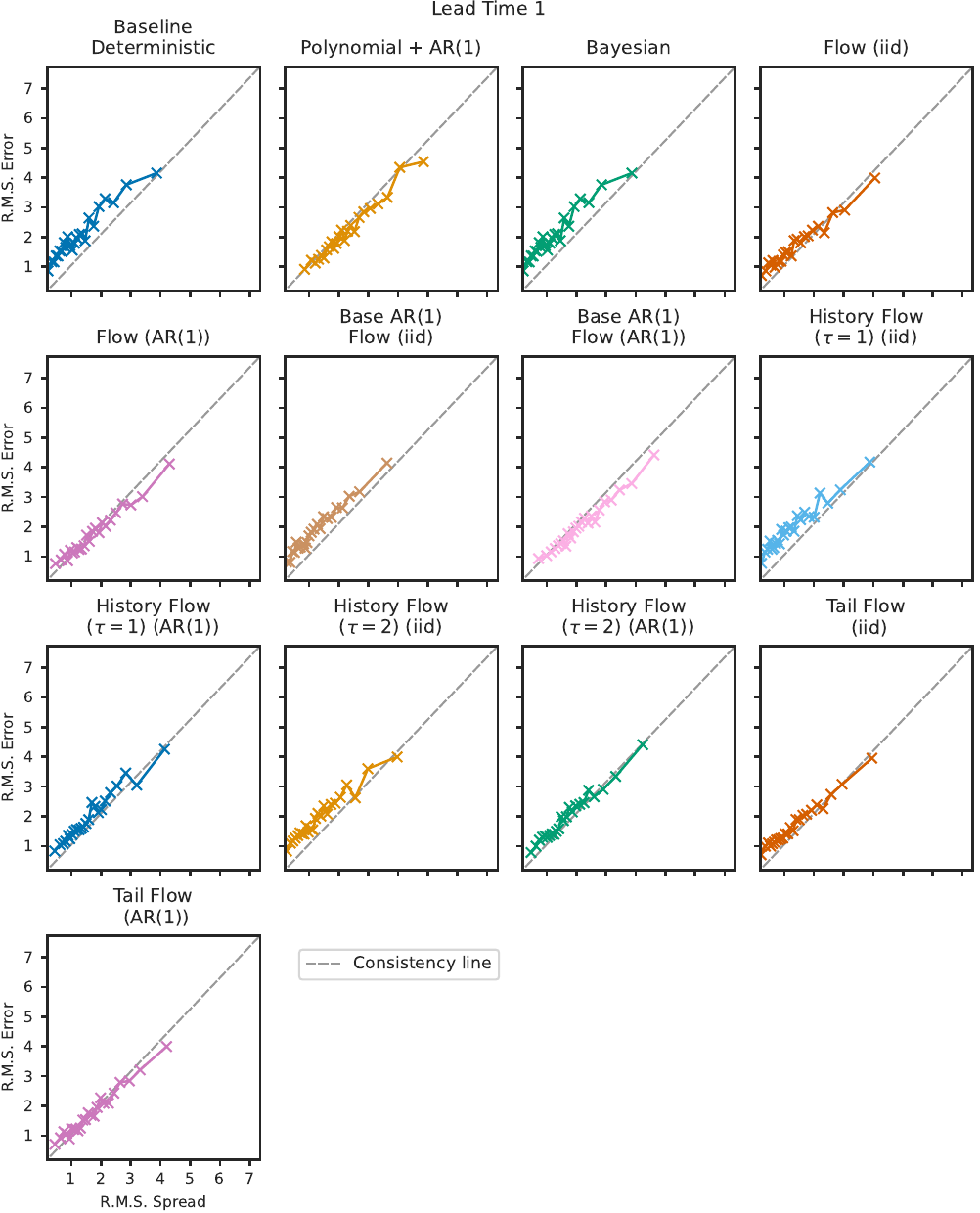}
    \caption{Spread-error consistency at lead time $\tau=1$ for the mixed ensemble configuration $(\ninit \! \times \! \nens)$. 
    This figure provides the mixed-ensemble counterpart to Fig.~\ref{fig:spread_error_grid_full_lt_1}. 
    Models above the consistency line are underdispersive, whereas models below it are overdispersive. 
    The results are consistent with the separated setting,  showing that the separated ensemble framework preserves the statistical consistency of the operationally used mixed setting.
    }
    \label{fig:spread_error_grid_mix_lt_1}
\end{figure}

\begin{figure}[ht]
    \centering
    \includegraphics[width=0.8\linewidth]{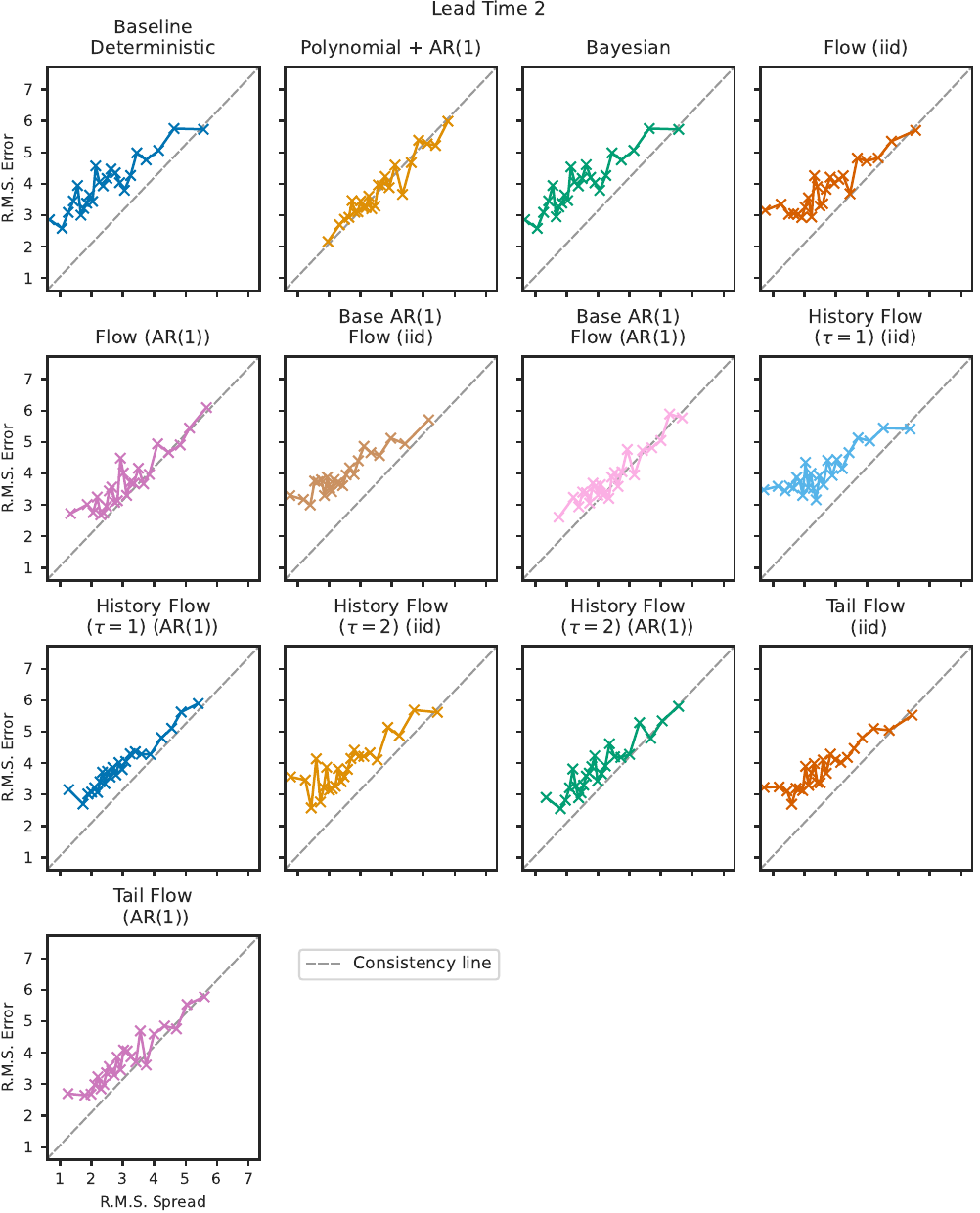}
    \caption{Spread-error consistency at lead time $\tau=2$ for the mixed ensemble configuration $(\ninit \! \times \! \nens)$. 
    This figure provides the mixed-ensemble counterpart to Fig.~\ref{fig:spread_error_grid_full_lt_2}. 
    Results remain consistent with the separated setting, with models generally becoming more underdispersive relative to $\tau=1$ (Fig.~\ref{fig:spread_error_grid_mix_lt_1}).
    }
    \label{fig:spread_error_grid_mix_lt_2}
\end{figure}

\begin{figure}[ht]
    \centering
    \includegraphics[width=0.8\linewidth]{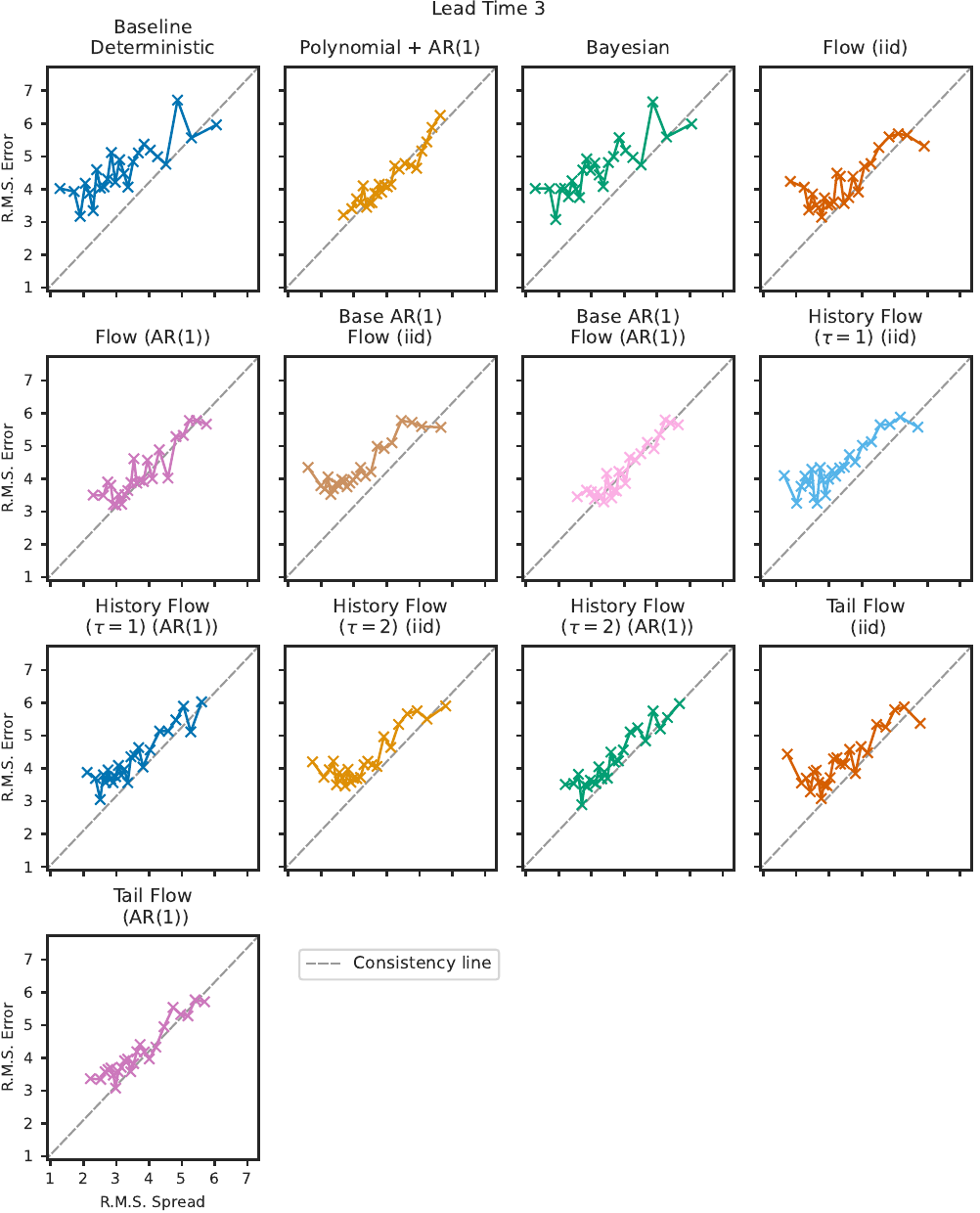}
    \caption{Spread-error consistency at lead time $\tau=3$ for the mixed ensemble configuration $(\ninit \! \times \! \nens)$. 
    This figure provides the mixed-ensemble counterpart to Fig.~\ref{fig:spread_error_grid_full_lt_3}. 
    Results are consistent with the separated setting, with only the best persistent models remaining close to the consistency line.
    }
    \label{fig:spread_error_grid_mix_lt_3}
\end{figure}

Since RMSE, ANCR, and spread-error diagnostics do not require separating perturbation- and model-induced contributions, we also evaluate them in the traditional mixed ensemble configuration $(\ninit \times \nens)$ with stochastic physics active (Appendix Sec.~\ref{sec:mixed_ensemble_appendix}).

In the mixed configuration, the ensemble mean is computed over ensemble members $j$ only, rather than jointly over ensemble members and stochastic realizations $(j,m)$. 
The corresponding expressions are obtained by replacing $X^{\mathrm{mean}_{j,m}}$ with $X^{\mathrm{mean}_{j}}$ in Eqs.~\eqref{eq:rmse_appendix}, \eqref{eq:a_mean_appendix}, \eqref{eq:spread_error_spread_appendix}, and \eqref{eq:spread_error_diff_appendix}. 
For the spread-error relationship, the ensemble size is $M=\nens$.

The mixed-configuration forecast metrics are shown in Fig.~\ref{fig:rmse_ancr_mix_appendix} and Tab.~\ref{tab:full_vs_mix_rmse_ancr_appendix} for RMSE and ANCR, and in Figs.~\ref{fig:spread_error_grid_mix_lt_1}, \ref{fig:spread_error_grid_mix_lt_2}, and \ref{fig:spread_error_grid_mix_lt_3} for spread-error diagnostics. 
All metrics are nearly identical between the separated configuration $(\ninit \! \times \! \nens \! \times \! \nmodel)$ and the mixed configuration $(\ninit \! \times \! \nens)$.
Thus, in terms of forecast skill and calibration, the separated ensemble framework is consistent with the traditional mixed setting used in operational practice.

\section{Correlation functions}

\begin{figure}
    \centering
    \includegraphics[width=0.8\linewidth]{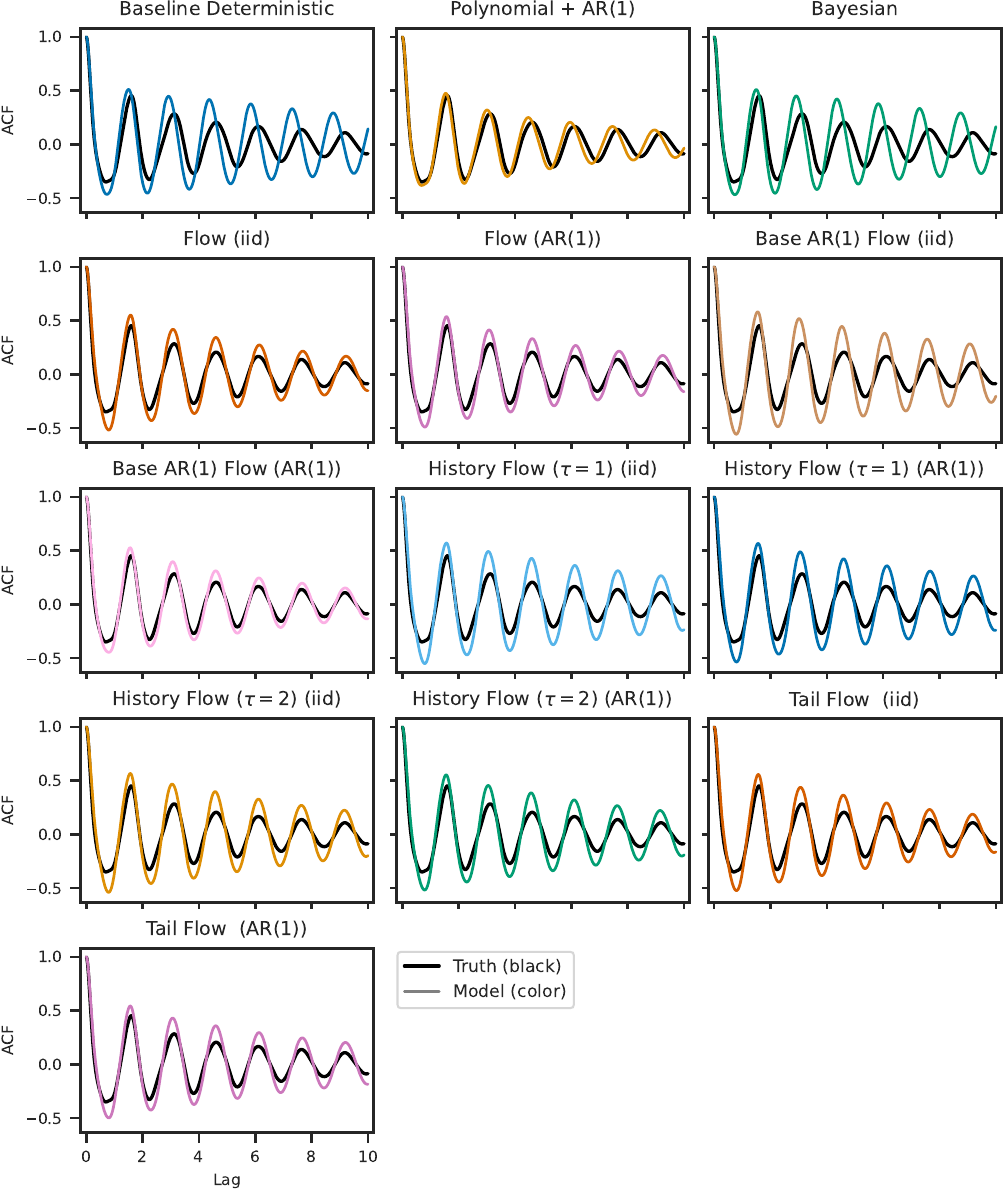}
    \caption{Autocorrelation function (ACF) averaged over all grid points $k$.
    Deterministic and Bayesian models lose phase coherence rapidly. 
    The polynomial+AR(1) model reproduces the correlation structure well at short lags (up to about $\tau=4$) but degrades at longer lags. 
    Flow-based models maintain phase coherence but overestimate amplitudes, indicating that correlations decay too slowly.}
    \label{fig:acf_avg_k_appendix}
\end{figure}

\begin{figure}
    \centering
    \includegraphics[width=0.8\linewidth]{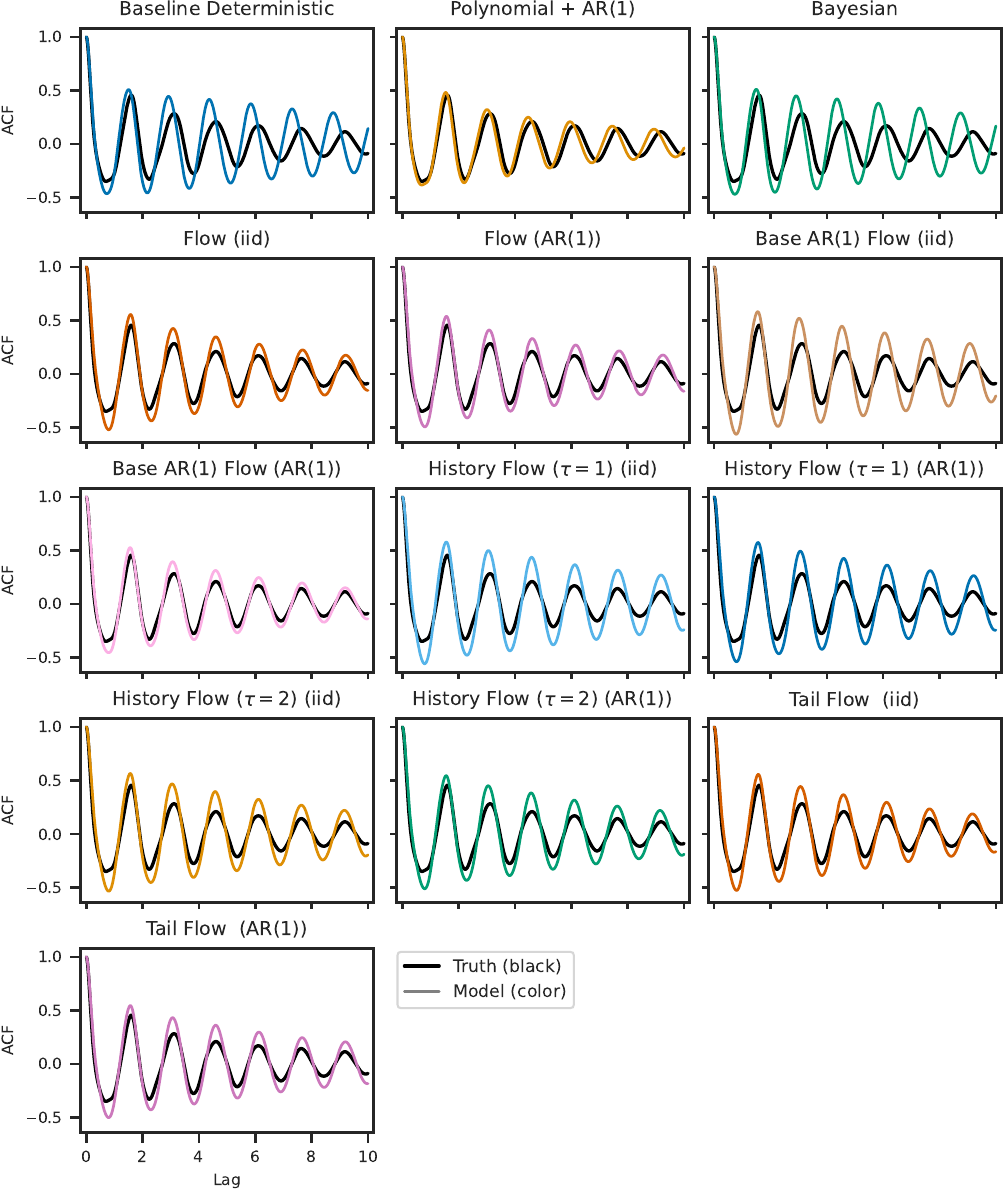}
    \caption{Autocorrelation function (ACF) for a single representative grid point $k=0$. 
    Results are nearly identical to the spatially averaged case shown in Fig.~\ref{fig:acf_avg_k_appendix}, demonstrating that averaging over $k$ does not obscure the temporal dynamics.}
    \label{fig:acf_k_0_appendix}
\end{figure}

\begin{figure}
    \centering
    \includegraphics[width=0.8\linewidth]{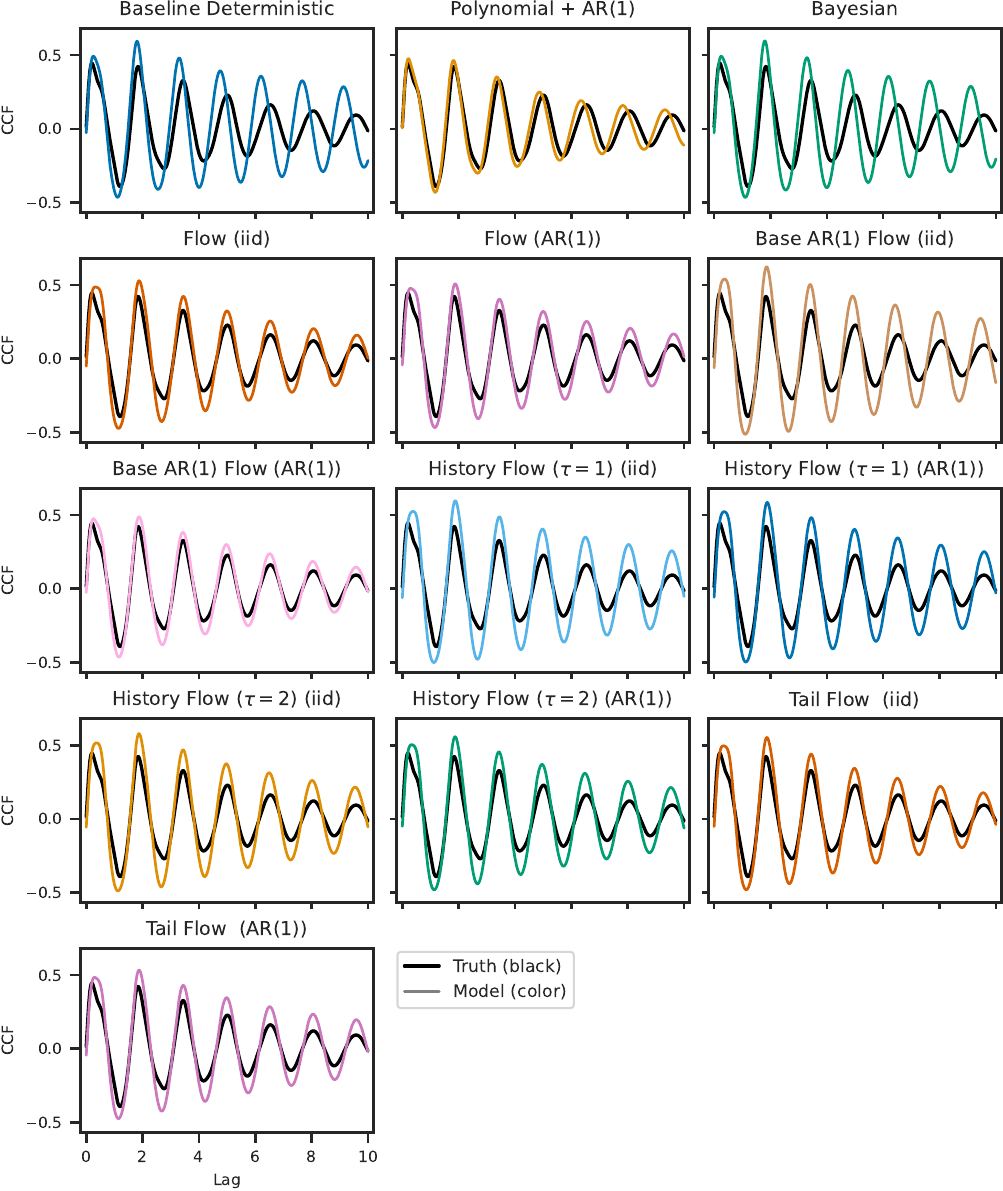}
    \caption{Cross-correlation function (CCF) between neighboring grid points $X_k$ and $X_{k+1}$, averaged over $k$. 
    Deterministic and Bayesian models fail to capture the phase relationship and lose coherence rapidly. 
    The polynomial + AR(1) model reproduces the propagation behavior well at short lags but develops phase errors at longer lags. 
    Flow-based models maintain phase coherence but overestimate amplitudes, indicating overly persistent correlations.}
    \label{fig:ccf_avg_k_appendix}
\end{figure}

\begin{figure}
    \centering
    \includegraphics[width=0.8\linewidth]{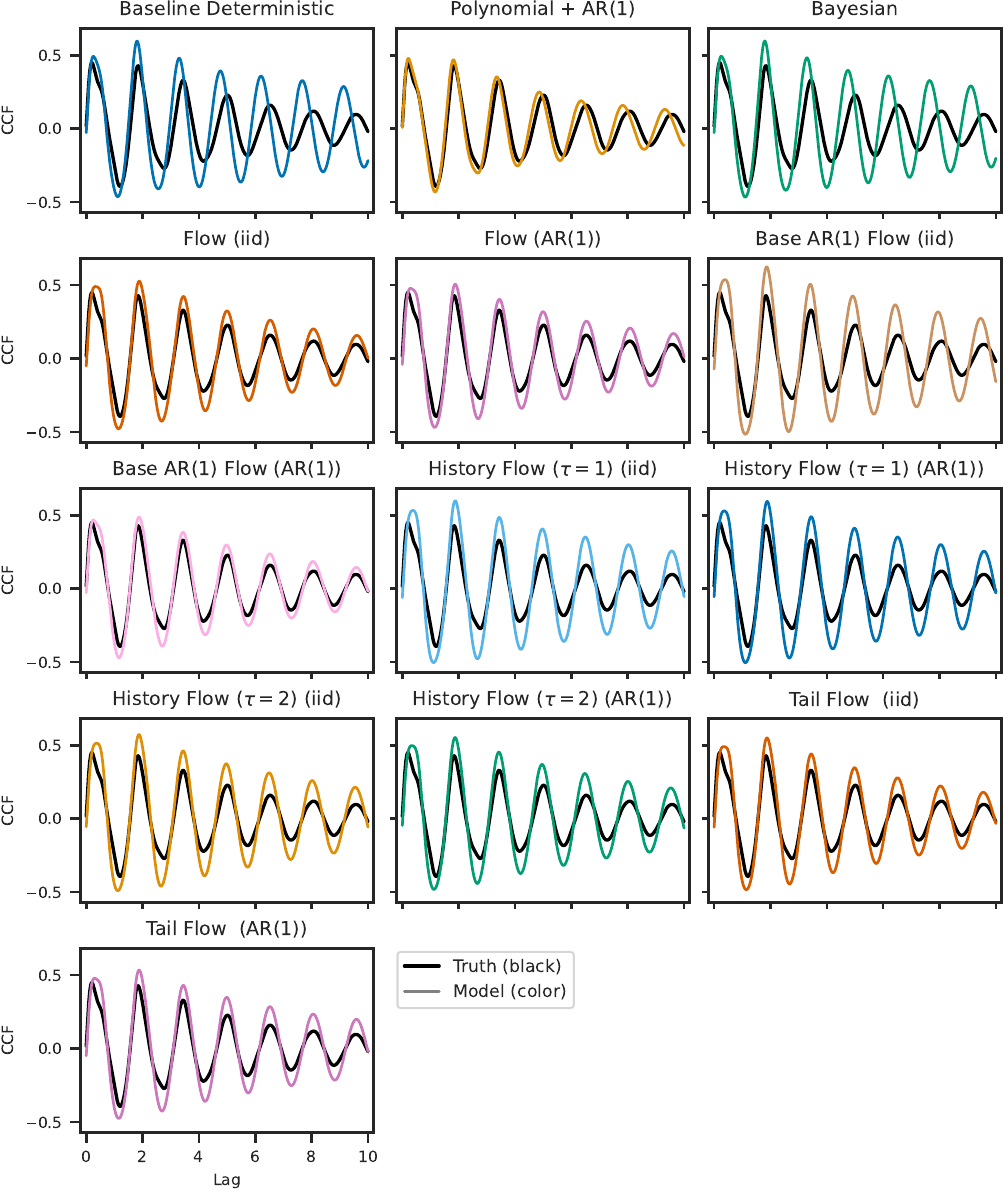}
    \caption{Cross-correlation function (CCF) between neighboring grid points $X_k$ and $X_{k+1}$ for a single representative grid point $k=0$. 
    Results are nearly identical to the spatially averaged case shown in Fig.~\ref{fig:ccf_avg_k_appendix}, demonstrating that averaging over $k$ does not obscure the temporal dynamics.}
    \label{fig:ccf_k_0_appendix}
\end{figure}

All forecast metrics in Sec.~\ref{sec:forecast_metrics_appendix} focus on the state-space distribution of the models. 
In contrast, autocorrelation (ACF) and cross-correlation (CCF) functions characterize the temporal and spatio-temporal structure of the dynamics in phase space.
The ACF measures the correlation of a time series with a time-shifted version of itself and thus quantifies temporal dependence across different lags, including persistence and oscillatory behavior. 
The CCF measures the correlation between two time series as a function of a time lag and captures lead-lag relationships. 
In the Lorenz '96 system, the CCF between neighboring grid points $X_k$ and $X_{k+1}$ is commonly used to characterize spatio-temporal propagation, such as phase shifts and wave-like dynamics.

For time series $X(t)$ and $Y(t)$, we define the ACF at lag $\tau$ as
\begin{align}
    acf(\tau) = \frac{Cov\left(X(t), X(t+\tau)\right)}{Var(X)},
\end{align}
and the CCF as
\begin{align}
    ccf(\tau) = \frac{Cov(X(t), Y(t+\tau))}{\sqrt{Var(X) \cdot Var(Y)}}.
\end{align}
Assuming stationarity, the mean and variance are independent of time, and the covariance depends only on the time lag $\tau$.

In practice, we estimate ACF and CCF from a finite trajectory of length $T$ using time averages:
\begin{align}
    \widehat{acf}(\tau) &= 
    \frac{\mathbb{E}
    \left[ 
    (X(t) - \mu_X) \cdot (X(t+\tau) - \mu_X)  
    \right]}
    {\mathbb{E}\left[ 
    (X(t) -\mu_X])^2
    \right]} \\
    &\approx
    \frac{\frac{1}{T-\tau}
    \sum_{t=0}^{T-\tau-1} \left( 
    (x(t) - \bar{x}) \cdot (x(t+\tau) - \bar{x})
    \right)
    }
    {\frac{1}{T}
    \sum_{t=0}^{T-1} (x(t) - \bar{x})^2
    }  
\end{align}
\begin{align}
    \widehat{ccf}(\tau) &= 
    \frac{
    \mathbb{E}
    \left[ 
    (X(t) - \mu_X) \cdot (Y(t+\tau) - \mu_Y)  
    \right]}
    {\sqrt{
    \mathbb{E}\left[  (X(t) - \mu_X)^2\right] 
    \cdot 
    \mathbb{E}\left[(Y(t) - \mu_Y)^2\right]  
    }} \\
    &\approx
    \frac{\frac{1}{T-\tau}
    \sum_{t=0}^{T-\tau-1} \left( 
    (x(t) - \bar{x})
    \cdot (y(t+\tau) - \bar{y})
    \right)
    }
    {
    \sqrt{\frac{1}{T}
    \sum_{t=0}^{T-1} (x(t) - \bar{x})^2
    \cdot 
    \frac{1}{T}
    \sum_{t=0}^{T-1}(y(t) - \bar{y})^2
    }
    }  
\end{align}
where $\mu_X = \mathbb{E}[X(t)]$ and $\mu_Y = \mathbb{E}[Y(t)]$ (stationarity) and  $\bar{x}$ and $\bar{y}$ denote sample means.

In our evaluation, we set $X = X_k$ and $Y = X_{k+1}$ and compute ACF and CCF from a long integration (10{,}000 MTU) up to lag $t=10$. 
Results are shown in Figs.~\ref{fig:acf_avg_k_appendix}--\ref{fig:ccf_k_0_appendix}  for both spatially averaged $k$ and a representative grid point $k=0$. 
The two cases are nearly identical, indicating that spatial averaging does not obscure the dynamics.

Across all models, stochastic parameterizations preserve temporal and spatio-temporal correlations more accurately than deterministic or Bayesian approaches, which lose phase coherence quickly. 
Flow-based models tend to overestimate correlation amplitudes, indicating overly persistent memory, while still capturing the correct oscillation frequency (no phase shift). 
This suggests that they reproduce the state dependence of the effective dynamics but retain memory for too long.
The polynomial+AR(1) model reproduces the true system behavior well up to approximately $t=4$, after which phase errors become visible in both ACF and CCF. 
Therefore, despite the simplicity of the polynomial+AR(1) parameterization, the temporal persistence improves short-lag correlations. 
Comparing the Base AR(1) flow with AR(1) sampling and the polynomial+AR(1) model, the simpler parameterizations better match the true amplitudes, while the flow model maintains phase coherence at longer lags.

\FloatBarrier
\section{Extended Figures}

\begin{figure}[th!]
    \centering
    \includegraphics[width=1\linewidth]{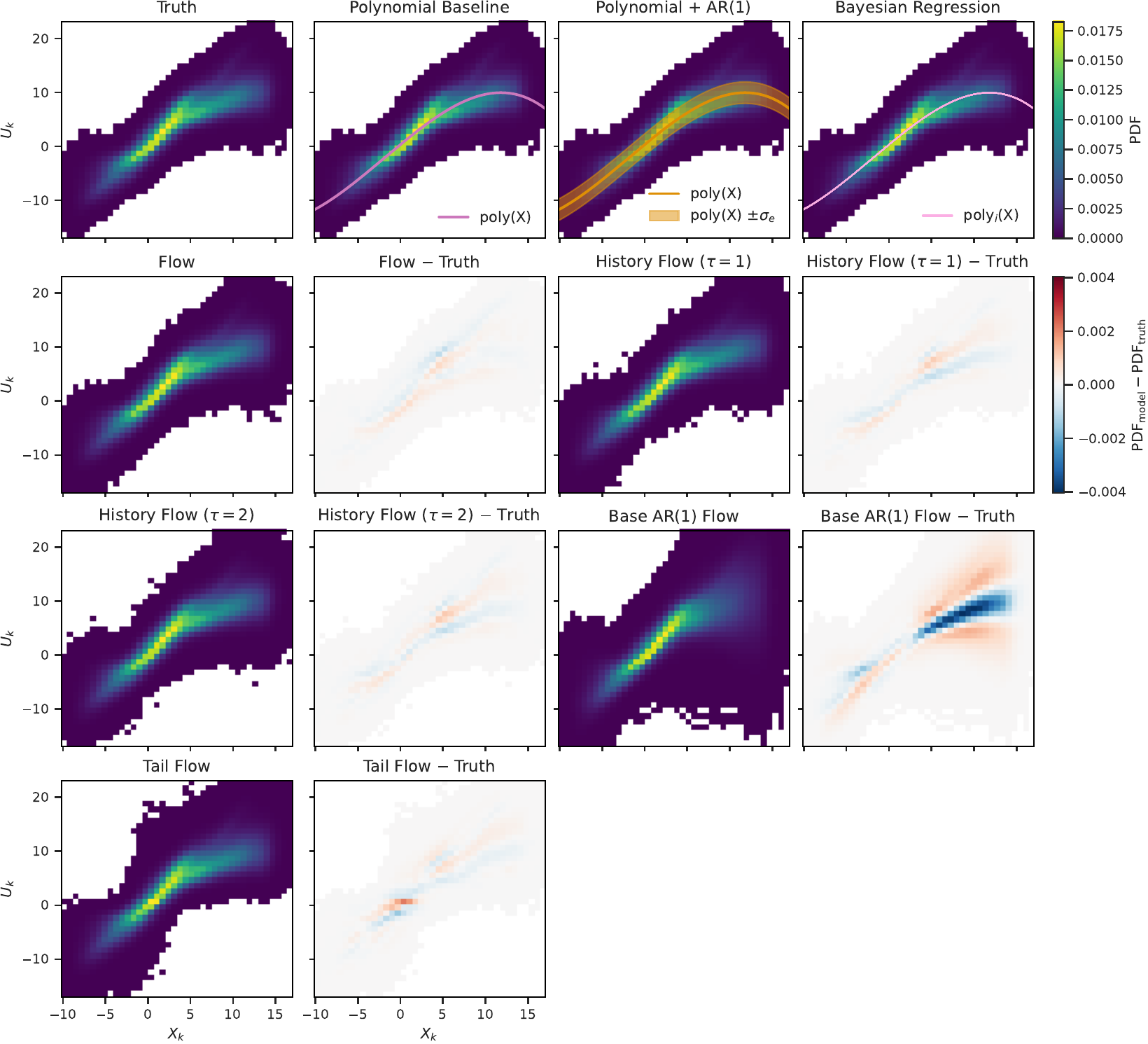}
    \caption{Extended comparison of joint distributions of $X_k$ and $U_k$.
    This figure supplements Fig.~\ref{fig:xu_fit} by including additional flow variants.
    All flow models sample $U_k$ conditionally on test states $X_k$ from the full system.}
    \label{fig:xu_fit_appendix}

\end{figure}

\begin{figure}
    \centering
    \includegraphics[width=0.8\linewidth]{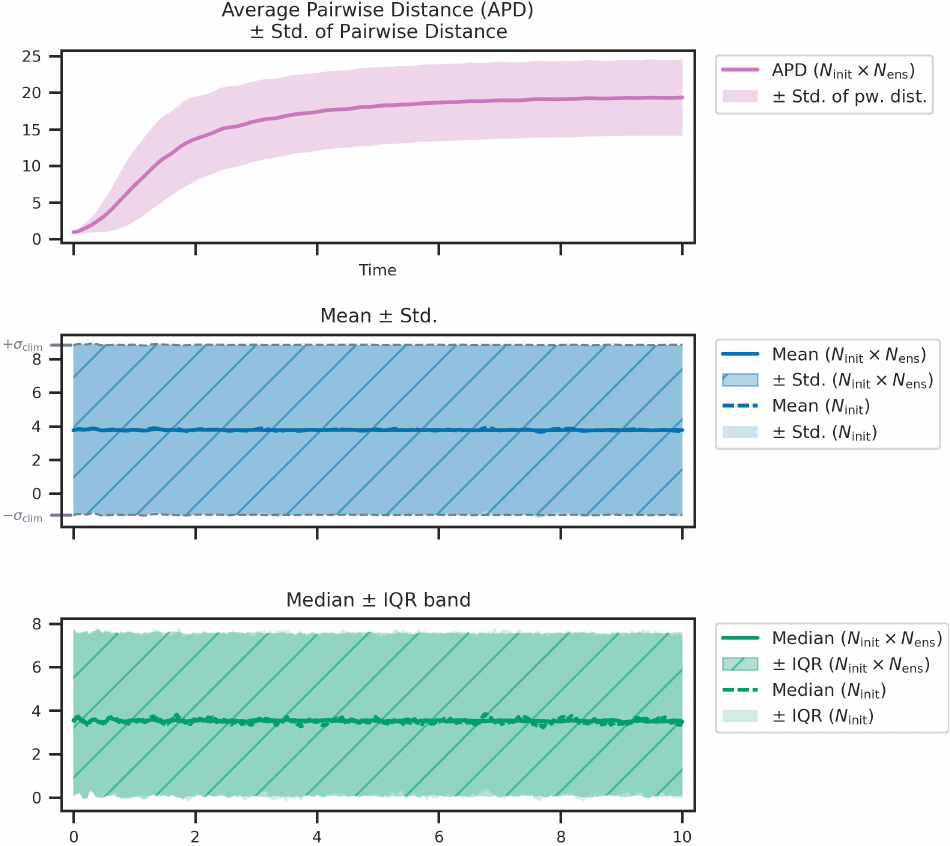}
    \caption{Additional predictability diagnostics of the fully resolved L96 system, averaged over spatial index $k$.
    The figure extends Fig.~\ref{fig:sensitivity} by reporting median and interquartile-range summaries in addition to the mean-based diagnostics.
    The robust summaries are consistent with the corresponding mean and variance diagnostics.}
    \label{fig:sensitivity_appendix_avg_k}
\end{figure}

\begin{figure}
    \centering
    \includegraphics[width=0.8\linewidth]{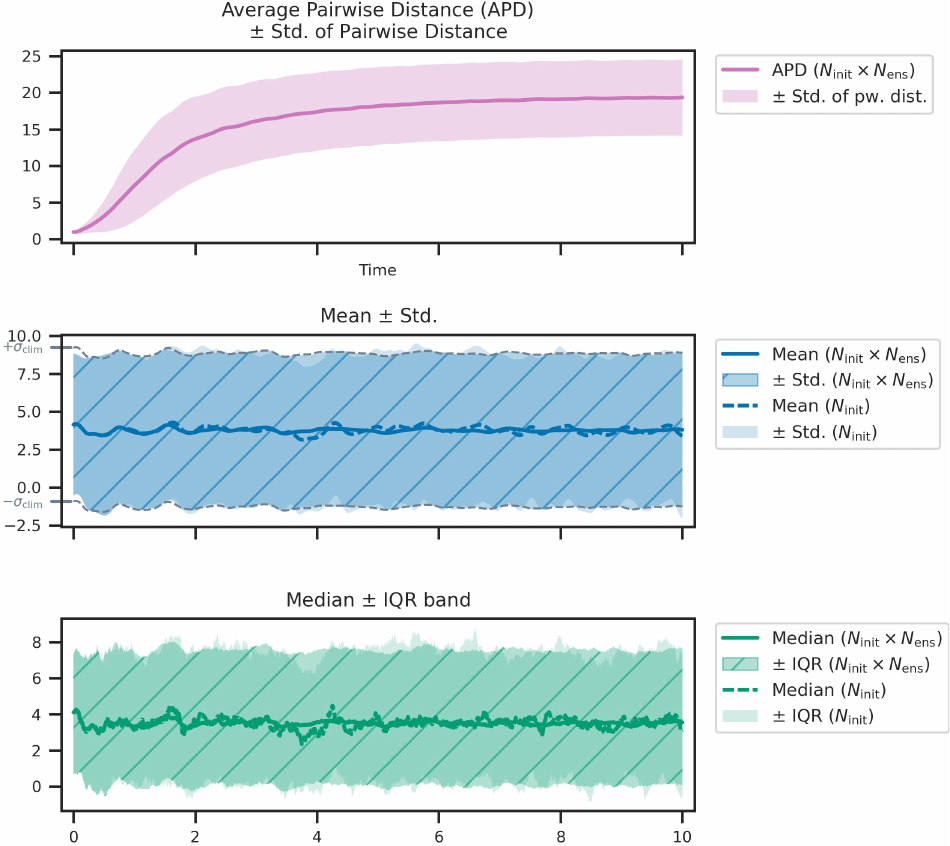}
    \caption{Predictability diagnostics of the fully resolved L96 system for spatial index $k=0$.
    This figure shows the same diagnostics as Fig.~\ref{fig:sensitivity_appendix_avg_k}, but for a single spatial index rather than averaged over $k$.
    The single-index diagnostics are noisier but remain consistent with the spatially averaged evaluation.
    Other spatial indices exhibit qualitatively similar results.}
    \label{fig:sensitivity_appendix_single_k}
\end{figure}

\begin{figure}
    \centering
    \includegraphics[width=0.8\linewidth]{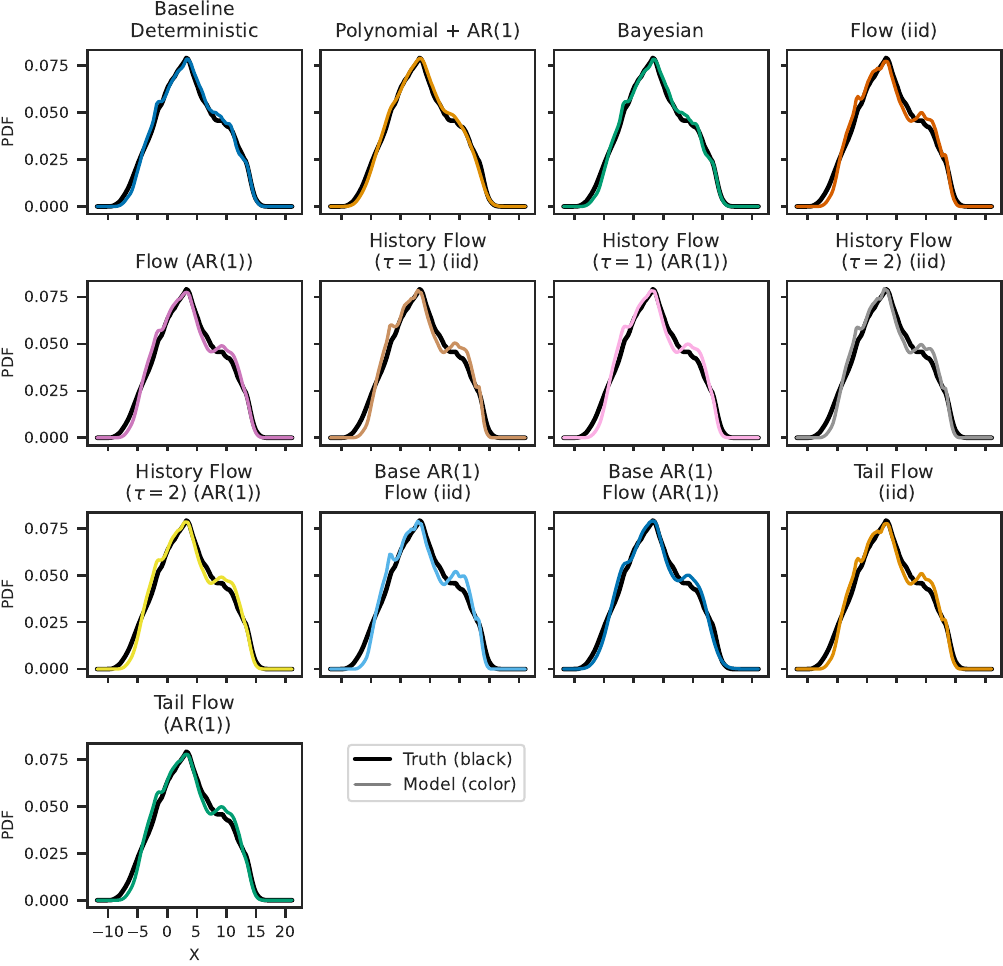}
    \caption{Marginal distributions of $X_k$ estimated from long integrations for all reduced models, compared to the true marginal distribution.
    This figure extends Fig.~\ref{fig:pdf} by including additional flow variants and i.i.d.\ sampling counterparts.
    AR(1) sampling improves the distributional fit relative to i.i.d. sampling, as quantified by the Hellinger distance and Kolmogorov-Smirnov statistic in Table~\ref{tab:hellinger_ks_appendix}.}
    \label{fig:pdf_appendix}
\end{figure}

\begin{figure}[t]
    \centering
    \includegraphics[width=0.8\linewidth]{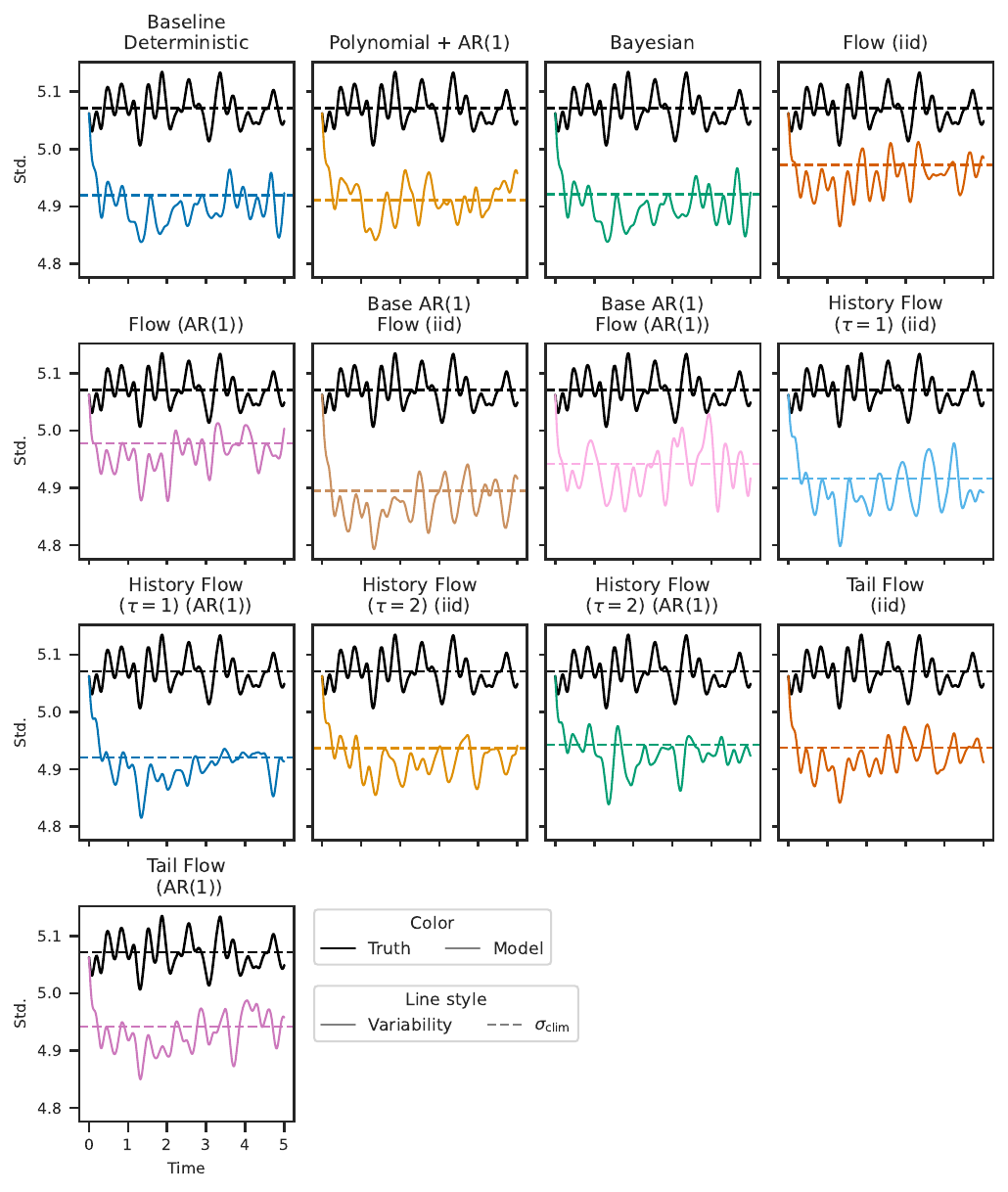}
    \caption{Time evolution of invariant-measure variability $\sigma^{\mathrm{init}}_k(t)$ together with the time-independent climatological amplitude $\sigma_{\mathrm{clim}}$ for the truth and all reduced models, averaged over spatial index $k$. 
    This figure extends Fig.~\ref{fig:iv} by including additional flow variants and i.i.d.\ sampling counterparts.
    In contrast to the marginal-distribution diagnostics in Fig.~\ref{fig:pdf_appendix}, AR(1) sampling does not improve the agreement of $\sigma_{\mathrm{clim}}$ with the truth compared with i.i.d. sampling.
    This suggests that temporal correlation mainly improves distributional structure rather than the overall climatological amplitude.}
    \label{fig:iv_avg_appendix}
\end{figure}

\begin{figure}[t]
    \centering
    \includegraphics[width=0.8\linewidth]{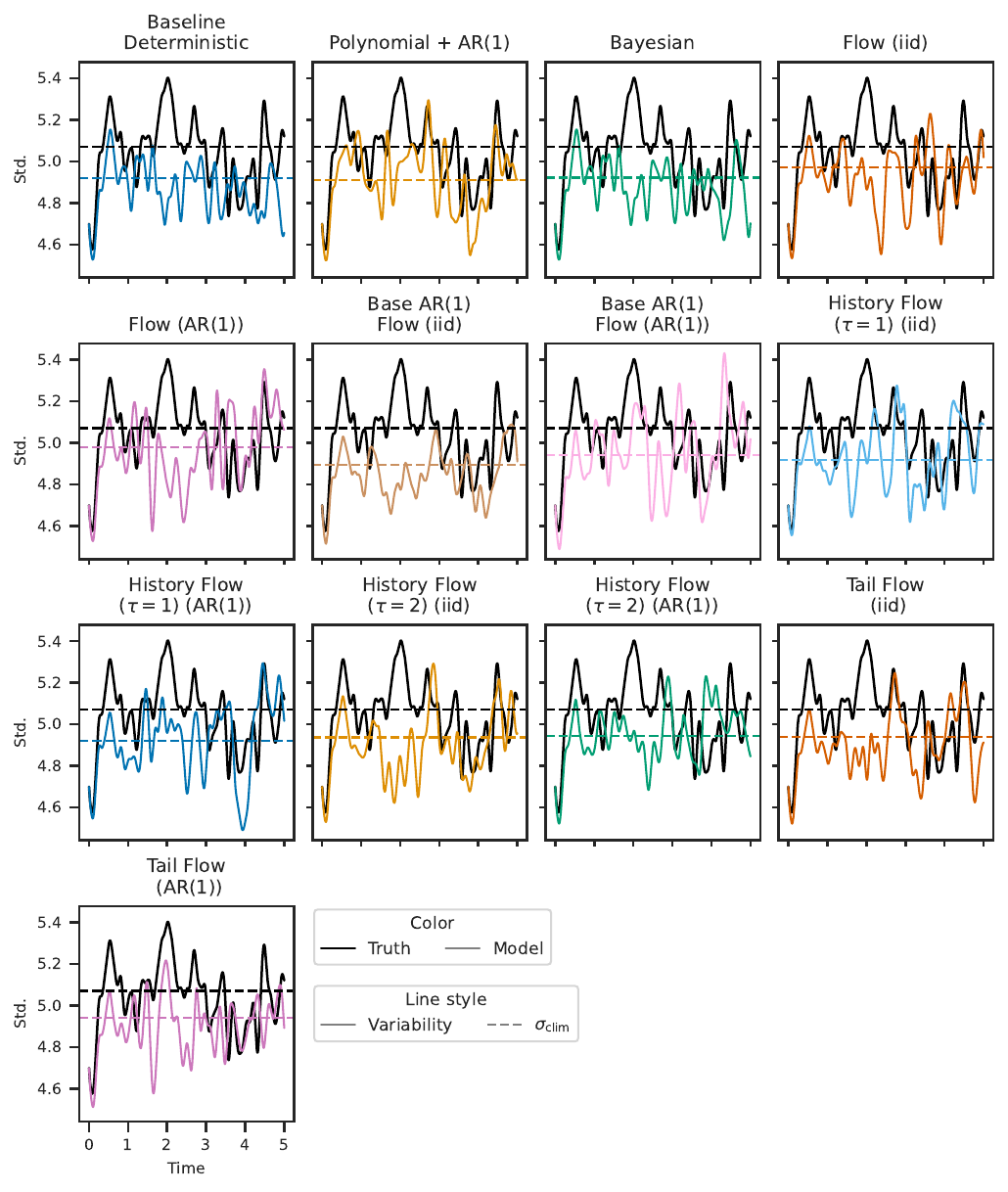}
    \caption{Time evolution of invariant-measure variability $\sigma^{\mathrm{init}}_k(t)$ together with the time-independent climatological amplitude $\sigma_{\mathrm{clim}}$ for the truth and all reduced models at spatial index $k=0$.
    This figure shows the same diagnostic as Fig.~\ref{fig:iv_avg_appendix}, but for a single spatial index rather than averaged over $k$.
    Single-$k$ results are noisier but show the same qualitative behavior as the spatially averaged results.
    Other spatial indices exhibit qualitatively similar behavior.}
    \label{fig:iv_k_appendix}
\end{figure}

\begin{figure}
    \centering
    \includegraphics[width=0.8\linewidth]{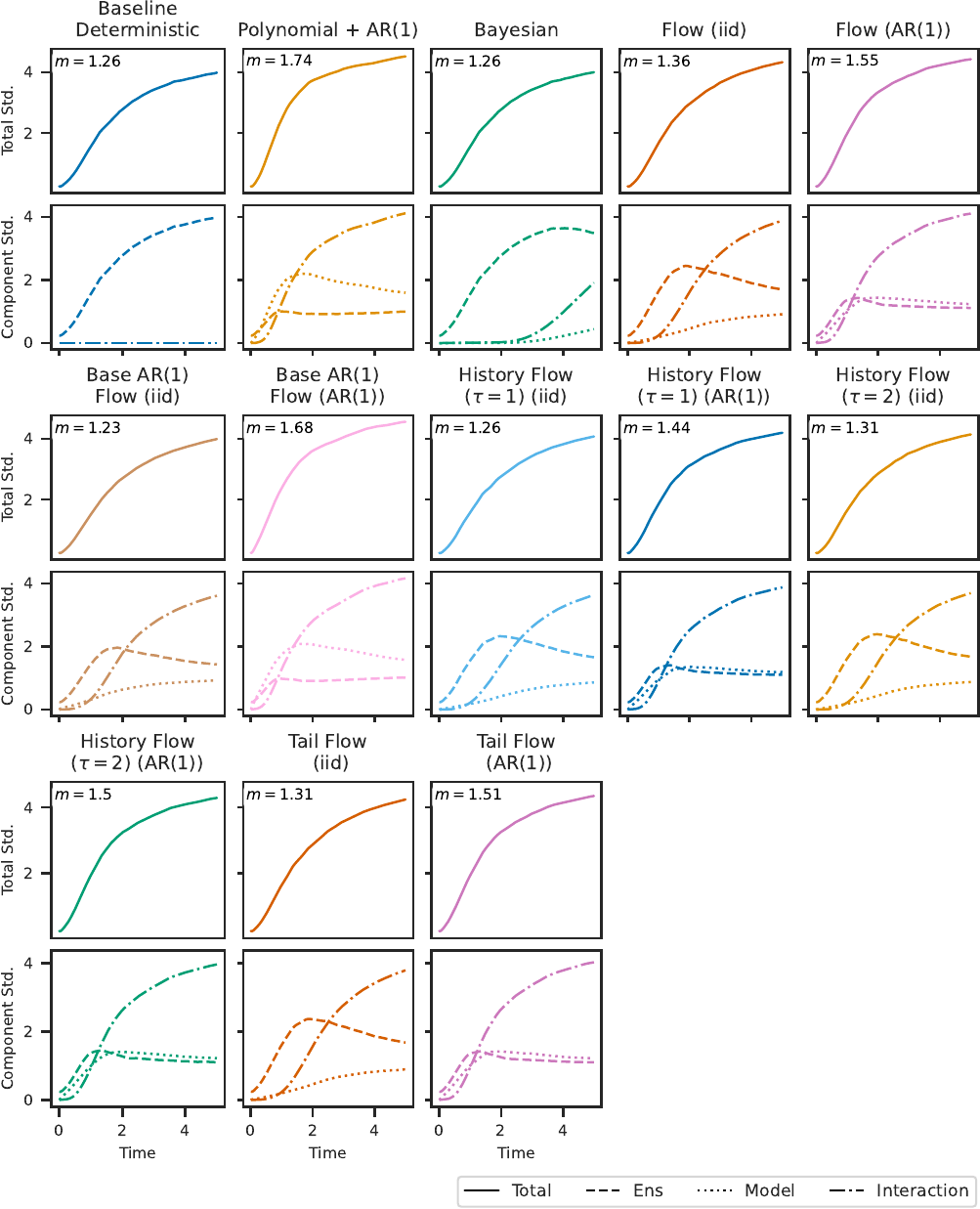}
    \caption{Decomposition of total initial-state-averaged ensemble spread into perturbation-, model-, and interaction-induced components, averaged over spatial index $k$.
    This figure extends Fig.~\ref{fig:spread_decomposition} by including additional flow variants and i.i.d.\ sampling counterparts.
    The value $m$ indicates the early-time growth rate of total spread, defined as the slope between $t=0$ and $t=2$.
    Compared with i.i.d.\ sampling, AR(1) sampling generally leads to an earlier increase in the model-induced component.
    This effect is strongest for the base AR(1) flow, where model-induced spread dominates the early growth.}
    \label{fig:spread_decomposition_avg_k_appendix}
\end{figure}

\begin{figure}
    \centering
    \includegraphics[width=0.8\linewidth]{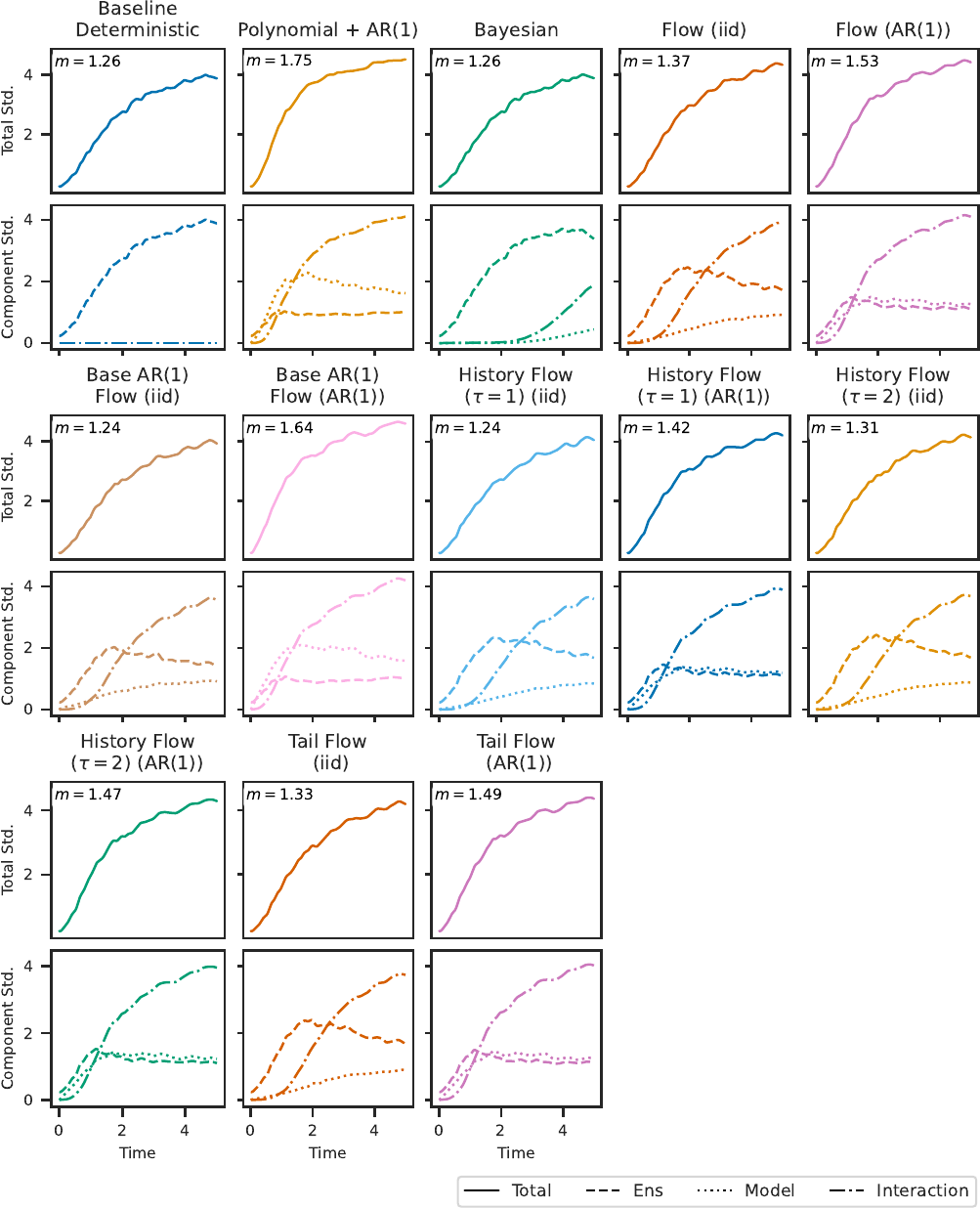}
    \caption{Decomposition of total initial-state-averaged ensemble spread into perturbation-, model-, and interaction-induced components for all models at spatial index $k=0$. 
    The value $m$ indicates the early-time growth rate of total spread, defined as the slope between $t=0$ and $t=2$.
    This figure shows the same diagnostic as Fig.~\ref{fig:spread_decomposition_avg_k_appendix}, but for a single spatial index rather than averaged over $k$.
    Single-$k$ results are noisier but show the same qualitative behavior as the spatially averaged results.
    Other spatial indices exhibit similar behavior.}
    \label{fig:spread_decomposition_k_appendix}
\end{figure}

\end{document}